\documentclass[10pt,letterpaper]{article}

\usepackage{arxiv}
\usepackage{graphicx}
\usepackage{amsmath}
\usepackage[title]{appendix}
\usepackage{amssymb}
\usepackage{booktabs}
\usepackage{xcolor}
\usepackage{multirow}
\usepackage{adjustbox}
\usepackage{xr}
\usepackage{floatrow}
\makeatletter
\newcommand*{\addFileDependency}[1]{
\typeout{(#1)}
\@addtofilelist{#1}
\IfFileExists{#1}{}{\typeout{No file #1.}}
}\makeatother

\usepackage[pagebackref,breaklinks,colorlinks]{hyperref}

\usepackage[capitalize]{cleveref}
\crefname{section}{Sec.}{Secs.}
\Crefname{section}{Section}{Sections}
\Crefname{table}{Table}{Tables}
\crefname{table}{Tab.}{Tabs.}

\usepackage[acronym]{glossaries}
\newacronym{aopc}{AoPC}{Area over Perturbation Curve}
\newacronym{rra}{RRA}{Relevance Rank Accuracy}
\newacronym{rma}{RMA}{Relevance Mass Accuracy}
\newacronym{ai}{AI}{Artificial Intelligence}
\newacronym{ml}{ML}{Machine Learning}
\newacronym{xai}{XAI}{Explainable AI}
\newacronym{bn}{BN}{BatchNorm}
\newacronym{vqa}{VQA}{Visual Question Answering}

\newacronym{lrp}{LRP}{Layer-wise Relevance Propagation}
\newacronym{ig}{IG}{Integrated Gradients}
\newacronym{gb}{GB}{Guided Backprop}
\newacronym{eb}{EB}{Excitation Backprop}
\newacronym{a2b1}{$\alpha$2$\beta$1}{Alpha2-Beta1}
\newacronym{ep}{$\varepsilon+$}{EpsilonPlus}

\renewcommand{\headeright}{}
\renewcommand{\undertitle}{}
\date{}
\begin{document}

\title{Optimizing Explanations by Network Canonization and Hyperparameter Search}

\author{%
    Frederik Pahde$^{1}$ \quad Galip Ümit Yolcu$^{1,2}$ \quad Alexander Binder$^{3,4}$ \quad Wojciech Samek$^{1,2,5}$ \quad Sebastian Lapuschkin$^1$\\
    $^1$Department of Artificial Intelligence, Fraunhofer Heinrich-Hertz-Institute\\$^2$Technische Universität Berlin\\ 
    $^3$ICT Cluster, Singapore Institute of Technology\\
    $^4$University of Oslo\\
    $^5$BIFOLD – Berlin Institute for the Foundations of Learning and Data \\
}

\maketitle

\begin{abstract}
Explainable AI (XAI) is slowly becoming a key component for many AI applications. 
Rule-based and modified backpropagation XAI approaches however often face challenges when being applied to modern model architectures including innovative layer building blocks, which is caused by two reasons.
Firstly, the high flexibility of rule-based XAI methods leads to numerous potential parameterizations. Secondly, 
many XAI methods break the implementation-invariance axiom because they struggle with certain model components, e.g., BatchNorm layers.
The latter can be addressed with model canonization, which is the process of re-structuring the model to disregard problematic components without changing the underlying function.
While model canonization is straightforward for simple architectures (e.g., VGG, ResNet), it can be challenging for more complex and highly interconnected models (e.g., DenseNet). 
Moreover, there is only little quantifiable evidence that model canonization is beneficial for XAI.
In this work, we propose canonizations for currently relevant model blocks applicable to popular deep neural network architectures, including VGG, ResNet, EfficientNet, DenseNets, as well as Relation Networks.
We further suggest a XAI evaluation framework with which we quantify and compare the effects of model canonization for various XAI methods in image classification tasks on the 
Pascal VOC and 
ILSVRC2017 datasets, 
as well as for Visual Question Answering using CLEVR-XAI. 
Moreover, addressing the former issue outlined above, we demonstrate how our evaluation framework can be applied to perform hyperparameter search for XAI methods to optimize the quality of explanations.
       
\end{abstract}

\section{Introduction}
\label{sec:intro}
In recent years, \gls*{ml} has been increasingly applied to high-stakes decision processes with a huge impact on human lives, such as medical applications~\cite{shahid_predictions_2020, brinker_deep_2019}, credit scoring~\cite{wang2011comparative}, criminal justice~\cite{zavrvsnik2020criminal}, and hiring decisions~\cite{bogen2018help}.
Therefore, awareness has been raised for the need of neural networks and their predictions to be transparent and explainable~\cite{goodman_european_2017}, which makes \gls*{xai} a key component of modern \gls*{ml} systems.
Rule-based  and modified backpropagation-based \gls*{xai} methods, such as DeepLift~\cite{shrikumar2017learning}, \gls*{lrp}~\cite{bach_pixel-wise_2015}, and Excitation Backprop~\cite{zhang_top-down_2018}, that are among the most prominent XAI approaches due to their high faithfulness and efficiency, however, struggle when being applied to modern model architectures with innovative building blocks. 
This is caused by two problems: 
Firstly, rule-based \gls*{xai}  methods provide large flexibility thanks to configurable rules which can be tailored to the model architecture at hand. 
This comes at the cost of a large number of potential \gls*{xai} method parameterizations, particularly for complex model architectures. 
However, finding optimal parameters is barely researched and often neglected, which can cause these methods to yield suboptimal explanations.
Secondly, earlier works~\cite{montavon2019gradient} have shown that many \gls*{xai} methods break implementation invariance, which has been defined as an axiom for explanations~\cite{sundararajan_axiomatic_2017}. 
This is caused by certain layer types for which no explanation rules have been defined yet, e.g., \gls*{bn} layers. 
To address that issue, \emph{model canonization} has been suggested, a method that fuses \gls*{bn} layers into neighboring linear layers without changing the underlying function of the model~\cite{hui_batchnorm_2019, guillemot_breaking_2020}, arguably leading to improved explanations for simple model architectures (VGG, ResNet)~\cite{motzkus_measurably_2022}. 
However, what constitutes a ``good'' explanation is only vaguely defined and many, partly contradicting, metrics for the quality of explanations have been proposed. 
Therefore, tuning hyperparameters of \gls*{xai} methods and measuring the benefits of model canonization for \gls*{xai} are non-trivial tasks.

To that end, we propose an evaluation framework, in which we evaluate \gls*{xai} methods w.r.t their faithfulness, complexity, robustness, localization capabilities, and behavior with regard to randomized logits, following the authors of~\cite{hedstrom_quantus_2022}. 
We apply our framework to (1) measure the impact of canonization and (2) demonstrate how hyperparameter search can improve the quality of explanations.
Therefore, we first extend the model canonization approach to modern model architectures with high interconnectivity, e.g., DenseNet variants. 
We apply our evaluation framework to measure the benefits of model canonization for various image classification model architectures (VGG, ResNet, EfficientNet, DenseNet) using the ILSVRC2017 and Pascal VOC 2012 
datasets, as well as for \gls*{vqa} with Relation Networks using the CLEVR-XAI dataset~\cite{arras_clevr-xai_2022}.
We show that generally model canonization is beneficial for all tested architectures, but depending on which aspect of explanation quality is measured, the impact of model canonization differs.
Moreover, we demonstrate how our \gls*{xai} evaluation framework can be leveraged for hyperparameter search to optimize the explanation quality from different points of view.

\section{Related Work}
\label{sec:related_work}

\subsection{XAI Methods}

\label{sec:rw_xai_methods}

XAI methods can broadly be categorized into local and global explanations. 
While local explainers focus on explaining the model decisions on specific inputs, global explanation methods aim to explain the model behavior in general, e.g., by visualizing learned representations.  
For image classification tasks, local XAI methods assign relevance scores to each input unit, expressing how influential that unit (e.g., an input pixel) has been for the inference process. 
Many XAI methods are (modified) backpropagation approaches. 
To compute the importance of features in the detection of a certain class 
, they start from the output of the network, backpropagating importance values layer by layer, depending on the parameters and/or hidden activations of each layer.
Saliency maps~\cite{simonyan2013deep,baehrens_saliency_2010,morch_saliency_1995}
are generated by computing the gradient $\frac{\partial f(\textbf{x}) }{\partial \textbf{x}}$, where $f(\textbf{x})$ is the model's prediction for an input sample $\textbf{x}$. This yields a feature map where each value indicates the model's sensitivity towards the corresponding feature.
Guided Backpropagation~\cite{springenberg_guided-bp_2014} also uses the gradients, but applies the ReLU function to computed gradients in ReLU activation layers in the backpropagation pass. 
This filters out the flow of negative information, allowing to focus on the parts of the image where the desired class is detected.
Integrated Gradients~\cite{sundararajan_axiomatic_2017} accumulates the activation gradients on a straight path in the input space, starting from a baseline image $\textbf{x}^\prime$ selected beforehand, to the datapoint of interest. Formally, the attribution to the $i^\text{th}$ feature is given by $(x_i-x^\prime_i)\int_{\rho=0}^1 \frac{\partial f(\textbf{x}+\rho(\textbf{x}^\prime -\textbf{x}))}{\partial x_i}d\rho$. 
SmoothGrad~\cite{smilkov_smooth_2017} aims to reduce the noise in saliency maps by sampling datapoints in the neighborhood of the original datapoint, and taking the average saliency map. 
Since these XAI methods rely only on the gradients of the total function computed by the network, they are implementation invariant, meaning they produce the same explanations for different implementations of the same function.

\gls*{lrp}~\cite{bach_pixel-wise_2015} operates by redistributing the relevance scores of neurons backwards up to the input features. 
More precisely, LRP distributes the activation of the output neuron of interest to the previous layers in a way that preserves relevance across layers. 
Several rules have been defined (e.g., LRP-$\epsilon$, LRP-$\gamma$, LRP-$\alpha\beta$), which can be combined in meaningful configurations according to the types and positions of layers in the neural network.
Excitation Backprop~\cite{zhang_top-down_2018} is a backpropagation method that is equivalent to LRP-$\alpha1\beta0$~\cite{montavon_methods-for-interpreting-dnns_2018}, which has a probabilistic interpretation. 
DeepLIFT~\cite{shrikumar_deeplift_2017} is another rule-based method, where a reference image (e.g., the mean over the training population) is selected in addition. Using the associated rules, the differences in the activations of neurons on the reference image and the target image are backpropagated to the input.
In addition to backpropagation-based XAI methods there are also other approaches.
Prominent examples are SHAP~\cite{lundberg_shap_2017}, which uses the game theoretic concept of Shapley values to find the contribution of each input feature to the model output, and  LIME~\cite{ribeiro_lime_2016}, which fits an interpretable model to the original model output around the given input. 
Both methods treat the model as a black box, only using outputs for certain inputs. As such, they are also implementation independent.  

\subsection{Evaluation of XAI Methods}
While various XAI methods have been developed, the quantitative evaluation thereof is often neglected and explanations of XAI methods are often only compared by visually inspecting heatmaps. 
To address this issue, many XAI metrics have been introduced in recent years~\cite{hedstrom_quantus_2022,agarwal2022openxai}. 
However, there is no consensus on which metric to use and moreover, each metric evaluates explanations from different viewpoints, partly with contradictory objectives. 
Broadly speaking, XAI metrics can be categorized into five classes: \textbf{Faithfulness} metrics measure whether an explanations truly represents features used by the model. For instance, Pixel Flipping~\cite{bach_pixel-wise_2015} measures the difference in output scores of the correct class, when replacing pixels in descending order of their relevance scores with a baseline value (e.g., black pixel or mean pixel). If the score decreases quickly, i.e., after replacing only a few highly relevant pixels, the explanation is considered as highly faithful. 
Region Perturbation~\cite{samek_evaluating_2016} further generalizes Pixel Flipping by replacing input regions instead of single pixels. Faithfulness correlation~\cite{bhatt_evaluating_2020} replaces a random subset of attribution with a baseline value and measures the correlation between the sum of attributions in the subset and the difference in model output.
\textbf{Robustness} metrics measure the robustness of explanations towards small changes in the input. Prominent examples are Max-Sensitivity and Avg-Sensitivity~\cite{yeh__2019}, which use Monte Carlo sampling to measure the maximum and average sensitivity of an explanation for a given XAI method.
\textbf{Localization} metrics measure how well an explanation localizes the object of interest for the underlying task. 
Consequently, in addition to the input sample and an explanation function, ground-truth localization annotations are required. Examples for localization metrics are \gls*{rra} and \gls*{rma}~\cite{arras_clevr-xai_2022}. 
\gls*{rra} measures the fraction of high-intensity relevances within the (binary) ground truth mask as $\text{RRA}=\frac{|P_{\text{top-K}} \cap GT|}{|GT|}$, where $GT$ is the ground truth, $K$ is the size of the ground truth mask and $P_{\text{top-K}}$ is the set of pixels sorted by relevance in decreasing order.
Similarly, \gls*{rma} measures the fraction of the total relevance mass within the ground truth mask and can be computed as $\text{RMA}=\frac{R_\text{within}}{R_\text{total}}$ where $R_\text{within}$ is the sum of relevance scores for pixels within the ground truth mask and $R_\text{total}$ is the sum of all relevance scores.
\textbf{Complexity} metrics measure how concise explanations are. For example, the authors of~\cite{chalasani_concise_2020} use the Gini Index of the total attribution vector to measure it's sparseness, while~\cite{bhatt_evaluating_2020} propose an entropy-derived measure.
\textbf{Randomization} metrics measure by how much explanations change when randomizing model components. 
For instance, the random logit test~\cite{sixt_when_2020} measures the distance between the original explanation and the explanation with respect to a random other class.


\subsection{Challenges of Rule-Based/Modified Backpropagation Methods}
\label{sec:xai_problems}
\paragraph{No Implementation Invariance:} From a functional perspective, it is desirable for an XAI method to be implementation invariant, i.e. the explanations for predictions of two different neural networks implementing the same mathematical function should always be identical~\cite{sundararajan_axiomatic_2017}. 
However rule-based and modified backpropagation approaches explain predictions from a message-passing point of view, which, by design, is affected by the structure of the predictor. 
This is impressively demonstrated by Montavon et al.~\cite{montavon2019gradient}, where the authors compute explanations for two different implementations of the same mathematical function and the relevance scores differ tremendously.
Therefore, these methods violate the implementation invariance axiom, for example because of concatenations of linear operations such as \gls*{bn} and Convolutional layers. 
However, this problem can be overcome with model canonization, i.e., re-structuring the network into a canonical form implementing exactly the same mathematical function.

\paragraph{Parameterization:} 
\label{sec:gamma_rule}
Rule-based backpropagation approaches are highly flexible and allow to tailor the XAI method to the underlying model and the task at hand. However, this flexibility comes at the cost of numerous different possible parameterizations.
For instance, the $\gamma$-rule in \gls*{lrp} computes relevances $R_j$ of layer $j$ given relevances $R_k$ from the succeeding layer $k$ as
\begin{equation}
    R_j=\sum_k\frac{a_j \cdot (w_{jk} + \gamma w_{jk}^{+})}{\sum_{j}a_j \cdot (w_{jk} + \gamma w_{jk}^{+})} \cdot R_k~~,
\end{equation}
where $a_j$ are the lower-layer activations, $w_{jk}$ are the weights between layers $j$ and $k$, $w_{jk}^{+}$ is the positive part of $w_{jk}$ and $\gamma$ is a parameter allowing to regulate the impact of positive and negative contributions. 
Therefore, $\gamma$ is a hyperparameter that has to be defined for each layer.
Note that the $\gamma$-rule becomes equivalent to the $\alpha1\beta0$-rule as $\gamma \rightarrow \infty$, where negative contributions are disregarded. 
Similarly, for $\gamma=0$, it is equivalent to the $\epsilon$-rule, where negative and positive contributions are treated equally. 
The choice of $\gamma$ for each layer can highly impact various measurable aspects of explanation quality.

\section{Model Canonization}
\label{sec:canonization_general}
We assume there is a model $f$, which, given input data $\textbf{x}$, implements the function $f(\textbf{x})$. 
We further assume that $f$ contains model components which pose challenges for the implementation of certain \gls*{xai} methods. 
Model canonization aims to replace $f$ by a model $g$ where $g(\textbf{x})=f(\textbf{x})$, but $g$ does not contain the problematic components. 
We call $g$ the canonical form of all models implementing the function $g(\textbf{x})$.
In practice, model canonization can be achieved by restructuring the model and combining several model components, as outlined in the following sections.
\subsection{BatchNorm Layer Canonization}
\label{sec:canonization_theory}
\gls*{bn} layers~\cite{ioffe_batch_2015} were introduced to increase 
the stability of model training by normalizing the gradient flows in neural networks.
Specifically, \gls*{bn} adjusts the mean and standard deviation as follows: 
\begin{equation}
    \text{BN}(\textbf{x})=w_{BN}^\top \Big(\frac{\textbf{x} - \mu}{\sqrt{\sigma + \epsilon}}\Big) + b_{BN}~~,
\end{equation}
where $w_{BN}$ and $b_{BN}$ are learnable weights and a bias term of the \gls*{bn} layer, $\mu$ and $\sigma$ are the running mean and running variance and $\epsilon$ is a stabilizer.
\newline
However, as discussed in Section \ref{sec:xai_problems}, \gls*{bn} layers have shown to pose challenges for modified backpropagation \gls*{xai} methods, such as \gls*{lrp}~\cite{hui_batchnorm_2019}. 
To address that problem, model canonization can be applied to remove \gls*{bn} layers without changing the output of the function.
We make use of the fact that during test time the \gls*{bn} operation can be viewed as a fixed affine transformation.
Specifically, we follow previous works~\cite{ guillemot_breaking_2020}, which have shown that \gls*{bn} layers can be fused with neighboring linear layers, including fully connected layers and Convolutional layers, of form  $w_L^\top \textbf{x} + b_L$, where $w_L$ is the weight matrix and $b_L$ is the bias term.
This results in a single linear layer, combining the affine transformations from the original linear layer and the \gls*{bn}.
The exact computation of the new parameters of the linear transformation depends on the order of model components:


\paragraph{Linear $\rightarrow$ BN:} Many popular architectures (including VGG~\cite{szegedy_going_2015} and ResNets~\cite{he_deep_2016}) apply batch normalization directly after Convolutional layers. Hence, this model component implements the following function:
\begin{align}
f(\textbf{x}) &= \text{BN}(\text{Linear}(\textbf{x})) \\
    &= (\underbrace{\frac{w_{BN}}{\sqrt{\sigma + \epsilon}} w_L}_{w_\text{new}})^\top \textbf{x} + \underbrace{\frac{w_{BN}}{\sqrt{\sigma + \epsilon}} (b_L - \mu) + b_{BN}}_{b_\text{new}}
    \label{eq:lin_bn}
\end{align}
which can be merged into a single linear layer with weight $w_{\text{new}}=\frac{w_{BN}}{\sqrt{\sigma + \epsilon}} w_L$ and bias $b_\text{new}=\frac{w_{BN}}{\sqrt{\sigma + \epsilon}} (b_L - \mu) + b_{BN}$. See Section~\ref{app:canonization_details} in the supplementary material for details.


\paragraph{BN $\rightarrow$ Linear:} Other implementations apply \gls*{bn} right before linear layers (i.e., \emph{after} the activation function of the previous layer), which impacts the computation of parameters of the merged linear transformation:
\begin{align}
f(\textbf{x}) &= \text{Linear}(\text{BN}(\textbf{x})) \\
&= \underbrace{\frac{w_L^\top w_{BN}}{\sqrt{\sigma + \epsilon}}}_{w_\text{new}} \textbf{x} \underbrace{-  \frac{w_L^\top w_{BN} \mu}{\sqrt{\sigma + \epsilon}} + w_L^\top b_{BN} + b_L}_{b_\text{new}} 
\label{eq:bn_lin}
\end{align} 
Again, this component can be fused into a single linear transformation with weight $w_{\text{new}}=\frac{w_L^\top w_{BN}}{\sqrt{\sigma + \epsilon}}$ and bias $b_\text{new}=w_L^\top b_{BN} - \frac{w_L^\top w_{BN} \mu}{\sqrt{\sigma + \epsilon}} + b_L$. 
Note that there are practical challenges when padding is applied, for instance in a Convolutional layer. 
In this case, the bias becomes a spatially varying term, which cannot be implemented with standard Convolutional layers. 
See Section~\ref{app:canonization_details} in the supplementary material for details.

\paragraph{BN $\rightarrow$ ReLU $\rightarrow$ Linear:} In some architectures, \gls*{bn} layers have to be merged with linear layers with an activation function (e.g., ReLU) in between. For instance, in DenseNets model components occur in that order. 
In that case, model canonization goes beyond merging two affine transformations, because of the non-linear activation function in between.
Therefore, we propose to swap the BN layer and the activation function, which can be achieved by defining a new activation function, named $\textit{ReLU}_{\textit{thresh}}$ which depends on the parameters of the \textbf{BN} layer, such that
\begin{align}
\text{ReLU}(\text{BN}(\textbf{x})) &= \text{BN}(\text{ReLU}_{\text{thresh}}(\textbf{x}))~~, 
\label{eq:bn_thresh}
\end{align}
where 
\begin{align}
\text{ReLU}_{\text{thresh}}(\textbf{x}) 
 = \begin{cases}
      \textbf{x} & \text{if } (w_{\text{BN}}>0 \text{ and } \textbf{x}>z)\\\textbf{x}&\text{if }(w_{\text{BN}}<0\text{ and } \textbf{x}<-z)\\
      z & \text{otherwise}
     \end{cases}
\label{eq:def_thresh}
\end{align}
with $z=\mu-\frac{b_{\text{BN}}}{w_{\text{BN}} / \sqrt{\sigma+\epsilon}}$.
Hence, \textbf{BN}$\rightarrow$\textbf{ReLU}$\rightarrow$ \textbf{Linear} is first transformed into \textbf{ThreshReLU}$\rightarrow$\textbf{BN}$\rightarrow$ \textbf{Linear}, then the \gls*{bn} layer and the linear layer can be merged with Eq.~\ref{eq:bn_lin}.
\begin{figure*}[ht]
\centering

  \includegraphics[width=.75\linewidth]{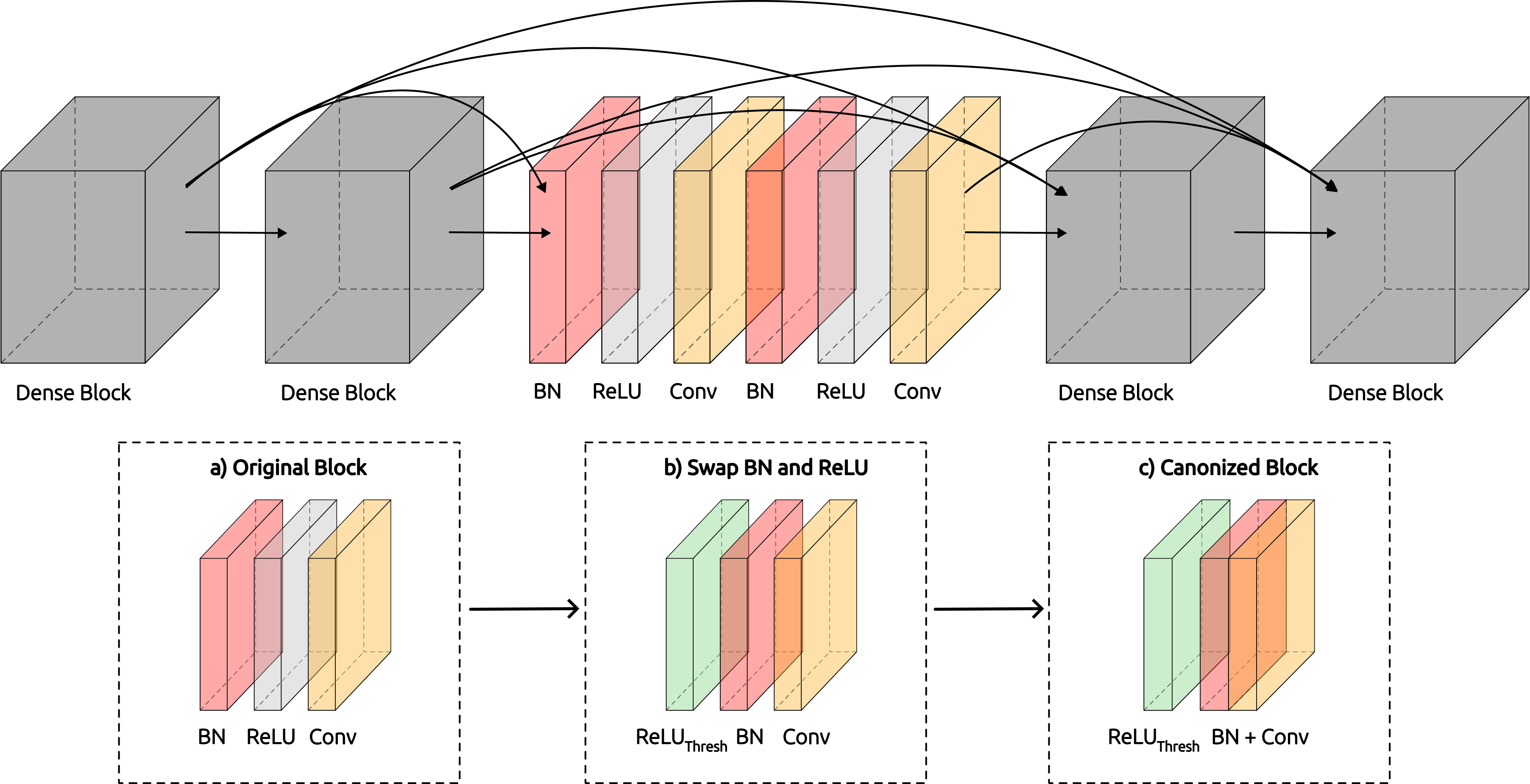}
  \caption{Due to the high interconnectivity of model components in DenseNets, it is not straightforward to fuse \gls*{bn} layers with Convolutional layers from neighboring blocks. Therefore, we suggest to first swap \gls*{bn} layers and ReLU activation functions using Eq.~\ref{eq:bn_thresh} (see step b)) and then merge the \gls*{bn} parameters into the following Convolutional layer using Eq.~\ref{eq:bn_lin} (see step c)).
  \label{fig:canon_densenet}}
\end{figure*}

\subsection{Canonization of Popular Architectures}
\label{sec:canonization_examples}
We now demonstrate the canonization of popular neural network architectures. 
We picked 4 image classification models (VGG, ResNet, EfficientNet and DenseNet) and one \gls*{vqa} model (Relation Network~\cite{santoro_simple_2017}). 

\paragraph{Image Classification Models:} Many popular image classification model architectures, such as VGG~\cite{szegedy_going_2015}, ResNet~\cite{he_deep_2016} and EfficientNet~\cite{tan_efficientnet_2019}, apply \gls*{bn} directly after linear layers. Therefore, these networks can easily be canonized using Eq.~\ref{eq:lin_bn}. 
It gets more complicated, however, if model architectures are more complex with highly interconnected building blocks. 
DenseNets, for example, use skip connections to pass activations from each dense block to all subsequent blocks, as shown in Fig.~\ref{fig:canon_densenet}. 
Each block applies \gls*{bn} on the concatenated inputs coming from multiple blocks, followed by ReLU activation and a Convolutional layer (BN $\rightarrow$ ReLU $\rightarrow$ Conv). 
Note that due to the high interconnectivity, \gls*{bn} layers cannot easily be merged into linear layers from neighboring blocks, because most linear layers pass their activations to multiple blocks, and vice versa, most blocks receive activations from multiple blocks.
Consequently, the linear transformation implementing the \gls*{bn} function has to be merged with the linear layer following the ReLU activation within the same block. 
Therefore, we propose to perform model canonization by first applying Eq.~\ref{eq:bn_thresh} and then Eq.~\ref{eq:bn_lin} to join \gls*{bn} layers with following linear layers within the same block, over the ReLU function between them.
This process is visualized in Fig.~\ref{fig:canon_densenet}.
In addition, we apply Eq.~\ref{eq:lin_bn} to merge the first \gls*{bn} layer in the initial layers of the DenseNet architecture, before the dense blocks. 
Moreover, there is a BN$\rightarrow$ReLU$\rightarrow$AvgPool2d$\rightarrow$Conv chain in the end of the network, which can also be merged using Eq.~\ref{eq:bn_thresh} and Eq.~\ref{eq:bn_lin}. 

\paragraph{VQA Model:} In contrast to image classification models, VQA models, e.g., Relation Network~\cite{santoro_simple_2017}, require two paths to encode both, the input image and the input question. Relation Networks use a simple Convolutional neural network as image encoder. The implementation by the authors of CLEVR-XAI~\cite{arras_clevr-xai_2022} applies \gls*{bn} \emph{after} the activation function (Conv $\rightarrow$ ReLU $\rightarrow$ BN). Therefore, we merge \gls*{bn} layers with the Convolutional layer from the following block using Eq.~\ref{eq:bn_lin}.
The \gls*{bn} layer of the last block of the image encoder has to be merged with the fully connected layer of the following module, which, however, receives a concatenation of image encoding and text encoding from the input question. 
Therefore, it has to be assured that \gls*{bn} parameters are merged only with the weights operating on inputs coming from the image encoder (see Section~\ref{sec:relation_network_details} and Fig.~\ref{fig:rn_canonization} in the supplementary material for details). 


\section{Experiments: XAI Evaluation Framework}
\label{sec:experiments}

\subsection{Datasets}
\label{sec:exp_datasets}

\paragraph{ILSVRC2017~\cite{russakovsky_imagenet_2015}} is a popular benchmark dataset for object recognition tasks with 1.2 million samples categorized into 1,000 classes, out of which we randomly picked 50 classes for our experiments (see Section~\ref{app:sec:imagenet_classes} in the supplementary material).
Bounding box annotations are provided for a subset of ILSVRC2017, which we use for localization metrics. Per class, we use up to 640 random samples. Note that ILSVRC2017 faces a center-bias, i.e., most of the objects to be classified are located in the center of the image. Therefore, naive explainers can assume that models base their decisions on center pixels. 
To that end, we include additional experiments using the Pascal VOC 2012 dataset~\cite{everingham_pascal_2010} in Section~\ref{sec:experiments_pascal}~in the supplementary material.


\paragraph{CLEVR-XAI~\cite{arras_clevr-xai_2022}} 
builds upon the CLEVR dataset~\cite{johnson_clevr_2017}, which is an artificial \gls*{vqa} dataset. It contains 10,000 images showing objects with varying characteristics regarding shape, size, color and material. 
Moreover, there are simple and complex questions that need to be answered. 
The task is framed as a classification task, in which, given an image and a question, the model has to predict the correct response out of 28 possible answers. 
In total, there are  approx.\ 40,000 simple questions, asking for certain characteristics of single objects. 
In addition, there are 100,000 complex questions, which require the understanding of relationships between multiple objects. 
CLEVR-XAI further comes with ground-truth explanations, encoded as binary masks locating the objects that are required in order to answer the question.
Simple questions come with two binary masks, which are \textit{GT Single Object}, localizing the object affected by the question, and \textit{GT All Objects}, localizing all objects in the image. 
For complex questions there are four binary masks, including \textit{GT Union} localizing all objects that are required to answer the question (we refer to~\cite{arras_clevr-xai_2022} for details on the other masks).



\subsection{Models}
\label{sec:exp_models}
For our experiments with ILSVRC2017, we analyze VGG-16~\cite{szegedy_going_2015}, ResNet-18~\cite{he_deep_2016}, EfficientNet-B0~\cite{tan_efficientnet_2019} and DenseNet-121~\cite{huang_densely_2017}. We use pre-trained models provided in the PyTorch model zoo~\cite{paszke_pytorch_2019}.
We use a Relation Network~\cite{santoro_simple_2017} for our experiments with CLEVR-XAI.

\subsection{XAI Methods and Implementation Details}
\label{sec:exp_explainers}
We analyze rule-based and modified backpropagation based XAI methods
, namely \gls*{eb} and \gls*{lrp}. 
Note that other backpropagation-based methods, such as Saliency, Smoothgrad, Integrated Gradients and Guided Backprop are not impacted by model canonization~\cite{motzkus_measurably_2022} and are therefore not analyzed in this experiment.
For each method, we compute explanations for both, the original and the canonized model. 
We use \textit{zennit}\footnote{\url{https://github.com/chr5tphr/zennit}}~\cite{anders_software_2021} as toolbox to compute explanations. 
For ILSVRC2017 with \gls*{lrp}, we analyze two pre-defined \emph{composites}, i.e., mappings from layer type to \gls*{lrp} rule which have been established in literature~\cite{montavon_layer-wise_2019}, namely \gls*{ep} and \gls*{a2b1}, see Tab.~\ref{tab:composites} in the supplementary  material for details. 
For Relation Networks, we use a custom composite (\emph{LRP-Custom}) following~\cite{arras_clevr-xai_2022}, in which we apply the $\alpha1\beta0$ rule to all linear layers and the box-rule~\cite{montavon_layer-wise_2019} to the input layer.
Note that ResNets, EfficientNets, and DenseNets leverage skip connections, which require the application of an additional canonizer in \textit{zennit} to explicitly make them visible to the XAI method.
Furthermore, we apply the signal-takes-it-all rule~\cite{arras__2017} to address the gate functions in the Squeeze-and-Excitation modules~\cite{hu_squeeze-and-excitation_2018} in EfficientNets.
In order to convert 3-dimensional relevance scores per voxel (channel $\times$ height $\times$ width) into 2-dimensional scores per pixel (height $\times$ width), we simply sum the relevances on the channel axis for ILSVRC2017 
experiments. 
For CLEVR-XAI experiments we follow the authors from~\cite{arras_clevr-xai_2022} and use 
\textit{pos-l2-norm-sq} ($R_\text{pool}=\sum_{i=1}^C max(0, R_i)^2$) as pooling function, where $C=3$ is the number of channels. 
Results for the alternative pooling function \textit{max-norm} ($R_\text{pool}=max(|R_1|, R_2, ..., R_C)$) are provided in the supplementary material.
Moreover, before computing the metrics, we normalize the relevances 
by dividing all values by the square root of the second moment to bound their variance for numerical stability when comparing heatmaps.

\subsection{XAI Metrics}
\label{sec:exp_metrics}
In our experiments, we quantitatively measure the impact of canonization of the selected model architectures with various metrics, probing the quality of explanations from different viewpoints. 
We use the \emph{quantus} toolbox\footnote{\url{https://github.com/understandable-machine-intelligence-lab/quantus}}~\cite{hedstrom_quantus_2022} to compute the following metrics:
We measure \textbf{Faithfulness} using Region Perturbation with blurring as baseline function. We compute the \gls*{aopc}~\cite{samek_evaluating_2016} to measure the faithfulness in a single number as $\text{AoPC}=\frac{1}{L+1}\Big(\sum_{k=0}^L f(\textbf{x}^{(0)}) - f(\textbf{x}^{(k)})\Big)$, where $\textbf{x}$ is the input sample, $k$ is the perturbation step and $L$ is the total number of perturbations. The \gls*{aopc} is averaged over all input samples. 
\textbf{Localization} quality is measured using \gls*{rra} and \gls*{rma}. As ground truth location we use bounding box annotations provided for ILSVRC2017, and binary segmentation masks for 
CLEVR-XAI. For the latter, we use \textit{GT Unique} for simple questions and \textit{GT Union} for complex questions. Moreover, we use the average sensitivity to measure \textbf{Robustness}, sparseness for \textbf{Complexity} and run the logit test as \textbf{Randomization} metric. While for robustness and randomization low scores are desirable, for the other metrics higher scores are better.

\begin{table*}[t]
    \centering
    \resizebox{\textwidth}{!}{ 
\begin{tabular}{l|l||cc|cc|cc|cc|cc|cc}
\toprule
{} & {} & \multicolumn{2}{l|}{$\uparrow$ Complexity} & \multicolumn{2}{l|}{$\uparrow$ Faithfulness} & \multicolumn{2}{l|}{$\uparrow$ Local. (RRA)} & \multicolumn{2}{l|}{$\uparrow$ Local. (RMA)} & \multicolumn{2}{l|}{$\downarrow$ Robustness} & \multicolumn{2}{l}{$\downarrow$ Random.} \\
Model & canonized &         no &   yes &           no &   yes &           no &   yes &           no &   yes &          no &   yes &            no &   yes \\
\midrule
\multirow{3}{*}{VGG-16} 

& EB       &       0.57 &  \textbf{0.59} &              0.35 &  \textbf{0.36} &         0.70 &  \textbf{0.71} &         0.68 &  \textbf{0.70} &        0.22 &  \textbf{0.18} &          1.00 &  1.00 \\
& LRP-$\alpha2\beta1$ &       0.70 &  \textbf{0.84} &              0.38 &  \textbf{0.39} &         0.63 &  \textbf{0.67} &         0.65 &  \textbf{0.77} &        \textbf{0.31} &  0.34 &          \textbf{0.59} &  0.66 \\
& LRP-$\varepsilon$+   &       0.51 &  \textbf{0.62} &              0.36 &  \textbf{0.39} &         0.69 &  \textbf{0.71} &         0.64 &  \textbf{0.71} &        \textbf{0.19} &  0.21 &          0.57 &  \textbf{0.54} \\

\midrule
\multirow{3}{*}{ResNet-18} 

& EB       &       0.55 &  \textbf{0.57} &              0.29 &  0.29 &         0.68 &  \textbf{0.69} &         0.66 &  \textbf{0.67} &        0.16 &  \textbf{0.14} &          0.97 &  0.97 \\
& LRP-$\alpha2\beta1$ &       0.67 &  \textbf{0.76} &              0.32 &  0.32 &         0.65 &  \textbf{0.67} &         0.69 &  \textbf{0.75} &        \textbf{0.21} &  0.26 &          0.65 &  \textbf{0.61} \\
& LRP-$\varepsilon$+   &       0.51 &  \textbf{0.58} &              0.30 &  0.30 &         0.69 &  \textbf{0.70} &         0.65 &  \textbf{0.69} &        \textbf{0.14} &  0.15 &          0.70 &  0.70 \\

\midrule
\multirow{3}{*}{EfficientNet-B0} 

& EB       &       \textbf{0.85} &  0.70 &              0.24 &  \textbf{0.27} &         \textbf{0.73} &  0.67 &         \textbf{0.79} &  0.72 &        0.42 &  \textbf{0.33} &          \textbf{0.99} &  1.00 \\
& LRP-$\alpha2\beta1$ &       0.75 &  \textbf{0.77} &              \textbf{0.29} &  0.20 &         \textbf{0.72} &  0.65 &         \textbf{0.79} &  0.73 &        \textbf{0.48} &  0.49 &          0.57 &  \textbf{0.51} \\
& LRP-$\varepsilon$+   &       0.50 &  \textbf{0.73} &              0.28 &  \textbf{0.30} &         0.75 &  0.75 &         0.69 &  \textbf{0.79} &        \textbf{0.12} &  0.21 &          \textbf{0.61} &  0.65 \\

\midrule
\multirow{3}{*}{DenseNet-121} 

& EB       &       \textbf{0.66} &  0.62 &              0.15 &  \textbf{0.31} &         0.58 &  \textbf{0.72} &         0.53 &  \textbf{0.73} &        0.57 &  \textbf{0.17} &          \textbf{0.75} &  0.89 \\
& LRP-$\alpha2\beta1$ &       \textbf{0.82} &  0.81 &              0.25 &  \textbf{0.33} &         0.64 &  \textbf{0.71} &         0.68 &  \textbf{0.81} &        0.65 &  \textbf{0.28} &          \textbf{0.40} &  0.44 \\
& LRP-$\varepsilon$+   &       \textbf{0.67} &  0.66 &              0.26 &  \textbf{0.33} &         0.70 &  \textbf{0.74} &         0.71 &  \textbf{0.77} &        0.63 &  \textbf{0.19} &          \textbf{0.39} &  0.48 \\

\bottomrule
\end{tabular}
}
            \caption{\label{tab:results_imagenet} \gls*{xai} evaluation results with and without model canonization for VGG-16, ResNet-18, EfficientNet-B0 and DenseNet-121 using the ILSVRC2017 dataset. We measure the quality of explanations using the metrics Sparseness (\emph{Complexity}), Region Perturbation (\emph{Faithfulness}), \gls*{rra} and \gls*{rma} (\emph{Localization}), Avg. Sensitivity (\emph{Robustness}) and Random Logit Test (\emph{Randomization}). Arrows indicate whether high ($\uparrow$) or low ($\downarrow$) are better. Best results are shown in bold.}
\end{table*}

\begin{table*}[ht]
    \centering
    
\resizebox{\textwidth}{!}{ 
\begin{tabular}{l|l||rr|rr|rr|rr|rr|rr}
\toprule
{}& {} & \multicolumn{2}{l|}{$\uparrow$ Complexity} & \multicolumn{2}{l|}{$\uparrow$ Faithfulness} & \multicolumn{2}{l|}{$\uparrow$ Local. (RRA)} & \multicolumn{2}{l|}{$\uparrow$ Local. (RMA)} & \multicolumn{2}{l|}{$\downarrow$ Robustness} & \multicolumn{2}{l}{$\downarrow$ Random.} \\
Questions & canonized &         no &   yes &           no &   yes &           no &   yes &           no &   yes &          no &   yes &            no &   yes \\
\midrule
\multirow{2}{*}{Simple} 
& EB                                     &       \textbf{0.99} &  0.97 &         0.50 &  \textbf{0.51} &         \textbf{0.64} &  0.61 &         \textbf{0.76} &  0.70 &        \textbf{1.37} &  1.39 &    1.00 &  1.0 \\
& LRP-Custom~\cite{arras_clevr-xai_2022} &       0.95 &  \textbf{0.98} &         0.52 &  0.52 &         0.70 &  0.70 &         0.75 &  \textbf{0.83} &        \textbf{1.33} &  1.35 &    \textbf{0.99} &  1.0 \\

\midrule
\multirow{2}{*}{Complex} 
& EB                                     &       \textbf{0.99} &  0.97 &         0.44 &  \textbf{0.45} &         \textbf{0.66} &  0.62 &         \textbf{0.82} &  0.77 &        1.36 &  \textbf{1.35} &    1.00 &  \textbf{0.99} \\
& LRP-Custom~\cite{arras_clevr-xai_2022} &       0.94 &  \textbf{0.97} &         0.45 &  \textbf{0.46} &         0.54 &  \textbf{0.63} &         0.79 &  \textbf{0.86} &        \textbf{1.33} &  1.34 &    \textbf{0.98} &  0.99 \\
\bottomrule
\end{tabular}
}
    \caption{\label{tab:results_clevr_xai} \gls*{xai} evaluation results for Relation Network with and without model canonization using the CLEVR-XAI dataset for simple and complex questions using \textit{pos-l2-norm-sq}-pooling. Arrows indicate whether high ($\uparrow$) or low ($\downarrow$) are better. Best results are shown in bold.}
\end{table*}

\subsection{Canonization Results}
\label{sec:exp_canon}
The \gls*{xai} evaluation results for the ILSVRC2017 dataset comparing models with and without canonization using the metrics described above are shown in Tab.~\ref{tab:results_imagenet}.
It can be seen that for most models and metrics the \gls*{xai} methods yield better explanations for canonized models, especially for complexity, faithfulness, and localization metrics.
However, there are some exceptions. 
While all other models yield less complex explanations when using model canonization, DenseNet explanations show the opposite behavior. 
This is due to the fact that many explanation heatmaps for DenseNets focus on small pixel groups (see Fig.~\ref{fig:attr_densenet_121} in supplementary material) that, however, do not truly represent the model's behavior, as low faithfulness scores without canonization indicate. 
The localization metrics tend to be better for canonized models, except for EfficientNet-B0 with the $\alpha2\beta1$-composite for \gls*{lrp}, which, however, also yields a poor performance in terms of faithfulness.
Hence, the $\alpha2\beta1$-composite itself appears to be a suboptimal parameterization for EfficientNets, which demonstrates the importance of the choice of hyperparameters for rule-based XAI methods.
Note that randomization scores tend to increase for canonized models, i.e., the canonized model leads to explanations that are less dependent on the target class. 
This is due to the fact that the explanations are more focused on the object to classify (see improvements for localization metrics in Tab.~\ref{tab:results_imagenet}) and therefore are more similar when computed for different target classes.
Hence, randomization metrics have to be interpreted with caution~\cite{binder2022shortcomings}.
Results for additional models and XAI evaluation metrics are shown in the supplementary material in Tables~\ref{tab:vgg16}-\ref{tab:densenet161}.

In Tab.~\ref{tab:results_clevr_xai} we show results for our XAI evaluation using the CLEVR-XAI dataset. 
For LRP-Custom, model canonization yields explanations that are either better than those of the original model or approximately\ on par with them for both simple and complex questions, in particular for localization metrics.
Results with \emph{max-norm}-pooling are shown in the supplementary material in Tab.~\ref{tab:clevr_max_norm}.

\label{sec:exp_results}

\begin{figure*}[ht]
\centering
  \includegraphics[width=.85\linewidth]{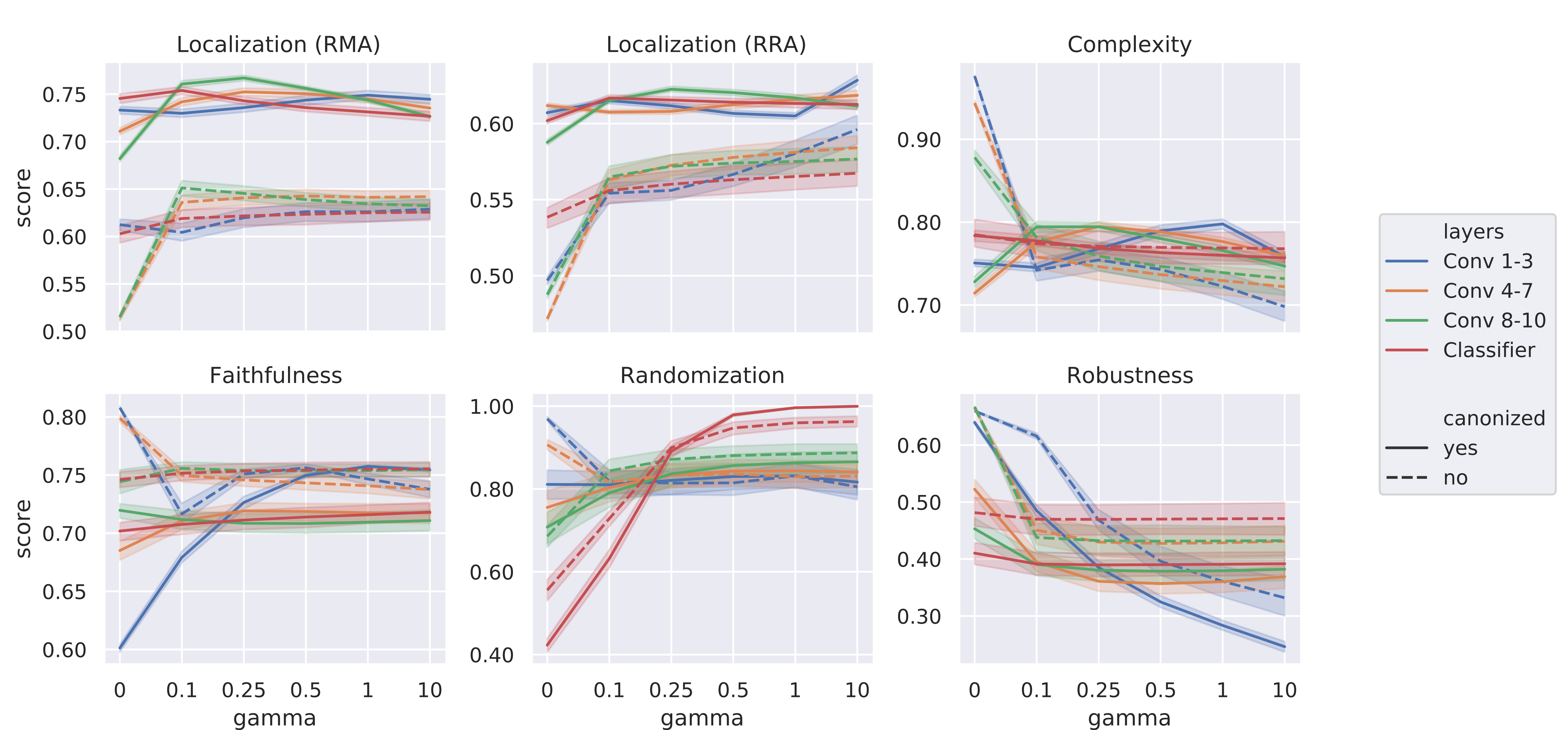}
  \caption{Results Grid Search for VGG-13: We group the layers into low-level (\emph{Conv 1-3}), mid-level (\emph{Conv 4-7}) and high-level (\emph{Conv 8-10}) Convolutions, as well as fully connected layers in the classification head (\emph{Classifier}). We evaluate different parameterizations for the $\gamma$-rule, where we define different values for $\gamma$ per group and measure the quality of explanations both with and without model canonization w.r.t localization, faithfulness, complexity, randomization and robustness metrics.}
  \label{fig:results_grid_search}
\end{figure*}

\begin{figure*}[ht]
\centering
  \includegraphics[width=.9\linewidth]{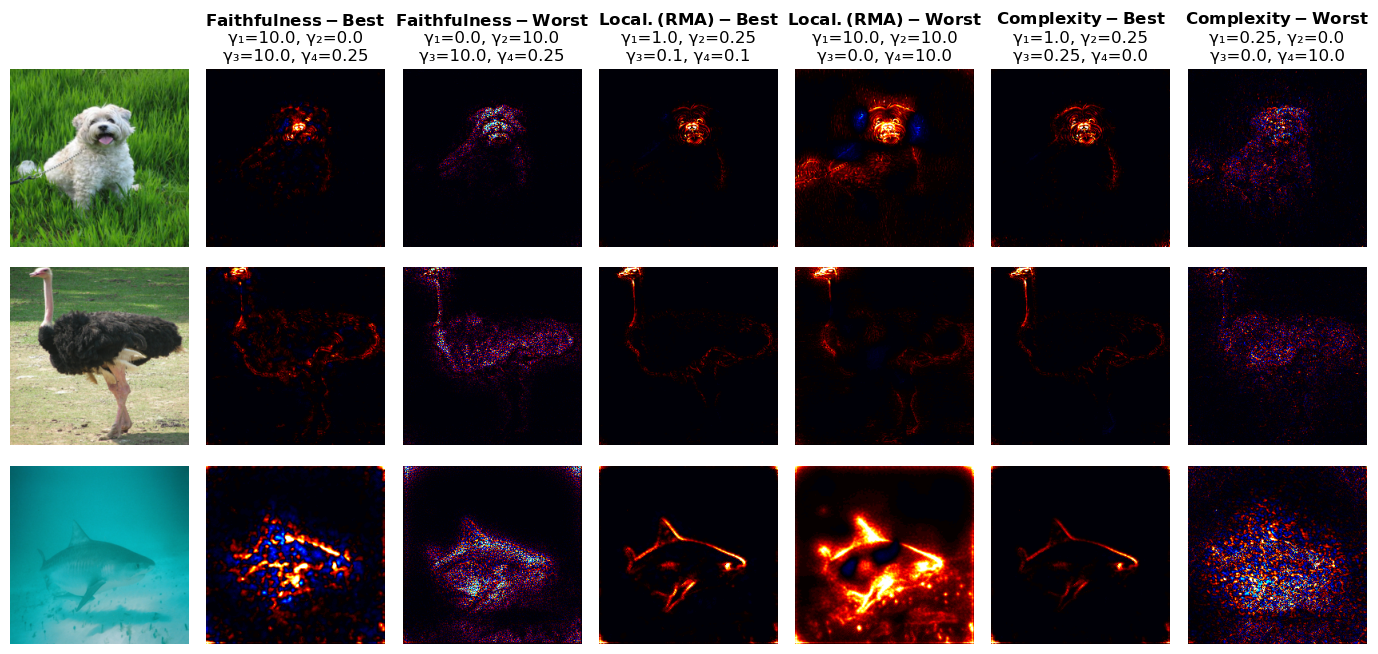}
  \caption{Attribution heatmaps (with canonization) for best and worst $\gamma$-parameters according to the grid search. $\gamma_1$ is for low-level features (Conv 1-4), $\gamma_2$ is for mid-level features (Conv 5-10), $\gamma_3$ is for high-level features (Conv 10-13), $\gamma_4$ is for layers in the classification head.  }
  \label{fig:attributions_grid_search}
\end{figure*}

\subsection{XAI Hyperparameter Tuning}
For our experiments in Section~\ref{sec:exp_canon} we used pre-defined LRP composites established in literature~\cite{montavon_layer-wise_2019,kohlbrenner_towards_2020}. 
However, as suggested by different results for evaluated composites, these parameters differently impact the quality of explanations w.r.t the chosen metric.
To that end, we run another experiment that uses our XAI evaluation framework for hyperparameter search.
Specifically, we focus on the LRP-$\gamma$-rule, which uses the parameter $\gamma$ to regulate the effect of positive and negative contributions (see Sec.~\ref{sec:gamma_rule}), ranging from treating both equally ($\gamma=0$) to neglecting negative contributions ($\gamma \rightarrow \infty$). 
Further, the flexibility of the \gls*{lrp} framework allows us to define \gls*{xai} methods with varying focus on positive and negative contributions depending on the position of the layer in the network.
We use a \mbox{VGG-13} model with \gls*{bn} and define 4 groups of network layers, which are low-level (Conv 1-3), mid-level (Conv 4-7), and high-level (Conv 8-10) layers, as well as fully-connected layers in the classification head. 
We define one $\gamma$-parameter per group with $\gamma \in \{0, 0.1, 0.25, 0.5, 1, 10\}$, and run a grid search for all possible combinations with and without model canonization, i.e., $2 \cdot 6^4=2592$ $\gamma$-configurations. 
Note that in theory, we could also evaluate different sub-canonizations, where we only canonize certain parts of the model.
This, however, further increases the degrees of freedom.
Further note that more advanced multi-metric-objective hyperparameter optimization approaches can be employed. 
However, we decided to go forward with simple grid search, because our goal is to highlight the importance of the choice of XAI hyperparameters and the impact on various evaluation metrics.
We evaluate the resulting explanations with the metrics described in section~\ref{sec:exp_metrics} and show the results in Fig.~\ref{fig:results_grid_search}. 
Specifically, each line represents the score per metric with $\gamma$ for a certain group of layers kept constant, averaged over all $\gamma$-parameterizations for other layer groups.
It can be seen that the impact of the choice of $\gamma$ depends on the position of the layer in the network and, in addition, differs by the metric of choice. 
For instance, the robustness of the explanations is mainly impacted by the $\gamma$-value in low-level layers (Conv 1-3), while it has no impact for the other layers.
In contrast, randomization is mostly impacted by the choice of $\gamma$ for fully connected layers in the classification head.  
Interestingly, canonization has a large impact on the optimal choice of $\gamma$ for low-level layers when measuring the faithfulness of the resulting explanations.
In Fig.~\ref{fig:attributions_grid_search} we show attribution heatmaps for three samples using different $\gamma$-configurations, employing the best and worst parameterization according to the metrics faithfulness, localization and complexity. 
Each metric favors another parametrization, leading to different attribution heatmaps. 
High $\gamma$-values in low-level layers ($\gamma_1$) appear to be favorable for all metrics, i.e., more focus on positive contributions on those layers.
This leads to attribution heatmaps with less noise, which is beneficial w.r.t faithfulness, localization and complexity.

\section{Conclusions}
In this work, we proposed an evaluation framework for XAI methods which can be leveraged to optimize the quality of explanations based on a variety of XAI metrics. 
Specifically, we demonstrated the application of our framework to measure the impact of model canonization towards various aspects of explanation quality. 
Therefore, we extended the model canonization approach to state-of-the-art model architectures, including EfficientNets and DenseNets.
Despite not always being beneficial w.r.t. all examined architectures, model canonization provides an extra option when adopting XAI methods to the task at hand.
Moreover, we applied our evaluation framework for hyperparameter optimization for XAI methods and demonstrated the impact of parameters w.r.t different XAI metrics.
While we have evaluated our methods for LRP, it is also applicable to other configurable XAI methods, such as DeepLift.
Future work will focus on the canonization of additional relevant model architectures, e.g., Vision Transformer~\cite{dosovitskiy2010image}.
In addition, optimizing the hyperparameter search is an promising research direction, e.g., with random search, evolutionary algorithms, or other approaches to reduce the search space. 
Moreover, the framework can be applied with other optimization objectives, e.g., to find LRP configurations that mimic other, more expensive XAI methods, e.g., SHAP.

\section*{Acknowledgements}
This work was supported by
the European Union's Horizon 2020 research and innovation programme (EU Horizon 2020) as grant [iToBoS (965221)];
the state of Berlin within the innovation support program ProFIT (IBB) as grant [BerDiBa (10174498)];
the Federal Ministry of Education and Research (BMBF) as grant [BIFOLD (01IS18025A, 01IS18037I)];
the Research Council of Norway, via the SFI Visual Intelligence grant [project grant number 309439].

{\small
\bibliographystyle{ieee_fullname}
\bibliography{egbib}
}
\clearpage
\begin{appendices}
\renewcommand{\thesection}{\Alph{section}}
\renewcommand\thefigure{A.\arabic{figure}}    
\renewcommand\thetable{A.\arabic{table}}
\renewcommand{\theequation}{A.\arabic{equation}}
\setcounter{figure}{0}   
\setcounter{table}{0}
\setcounter{equation}{0}

\section{Canonization Details}
\label{app:canonization_details}

\paragraph{Linear $\rightarrow$ BN:} In Eqs.~\eqref{appeq:details_canon_lin_bn_begin}~-~\eqref{appeq:details_canon_lin_bn_end}, we show more detailed steps required to fuse Linear $\rightarrow$ BN components into a single affine transformation, as outlined in Eq.~(4) in the main paper:
\begin{align}
\label{appeq:details_canon_lin_bn_begin}
f(\textbf{x}) &= \text{BN}(\text{Linear}(\textbf{x})) \\
    &= \text{BN}(w_L^\top \textbf{x} + b_L) \\
    &= w_{BN} \Big( \frac{w_L^\top \textbf{x} + b_L - \mu}{\sqrt{\sigma + \epsilon}}\Big) + b_{BN} \\
    &= \frac{w_{BN}}{\sqrt{\sigma + \epsilon}} (w_L^\top \textbf{x} + b_L - \mu) + b_{BN} \\
    &= (\underbrace{\frac{w_{BN}}{\sqrt{\sigma + \epsilon}} w_L}_{w_\text{new}})^\top \textbf{x} + \underbrace{\frac{w_{BN}}{\sqrt{\sigma + \epsilon}} (b_L - \mu) + b_{BN}}_{b_\text{new}}
    \label{appeq:details_canon_lin_bn_end}
\end{align}

\paragraph{BN $\rightarrow$ Linear:} In Eqs.~\eqref{appeq:can_bn_lin_begin}~-~\eqref{appeq:can_bn_lin_end}, we show more detailed steps required to fuse BN $\rightarrow$ Linear component chains into a single affine transformation, as outlined in Eq.~(6) in the main paper:
\begin{align}
f(\textbf{x}) &= \text{Linear}(\text{BN}(\textbf{x})) \label{appeq:can_bn_lin_begin} \\
&= w_L^\top \Bigg( w_{BN} \Big( \frac{\textbf{x} - \mu}{\sqrt{\sigma + \epsilon}}\Big) + b_{BN} \Bigg) + b_L \\
&= w_L^\top \Big( \frac{w_{BN} \textbf{x} - w_{BN} \mu}{\sqrt{\sigma + \epsilon}} + b_{BN} \Big) + b_L \\
&= \underbrace{\frac{w_L^\top w_{BN}}{\sqrt{\sigma + \epsilon}}}_{w_\text{new}} \textbf{x} \underbrace{-  \frac{w_L^\top w_{BN} \mu}{\sqrt{\sigma + \epsilon}} + w_L^\top b_{BN} + b_L}_{b_\text{new}} 
\label{appeq:can_bn_lin_end}
\end{align} 

 \textbf{Padding Issue in BN $\rightarrow$ Linear Canonization:} If the linear layer is a Convolutional layer with constant valued padding, the bias of the linear layer after canonization can no longer be shown as a scalar: 
\begin{align}
\hspace{-0.7cm}f(\textbf{x})&=\text{Conv$($Pad$($BN$($\textbf{x}})))\\
&=\text{Conv$($Pad$($}\frac{w_{BN}}{\sqrt{\sigma+\epsilon}}\textbf{x}-\frac{w_{BN}\mu}{\sqrt{\mu+\epsilon}}+b_{BN}))\\
&=\text{Conv(}\frac{w_{BN}}{\sqrt{\sigma+\epsilon}}\text{Pad}(\textbf{x})+\text{Pad}(-\frac{w_{BN}\mu}{\sqrt{\sigma+\epsilon}}+b_{BN}))\\
&=w_{L}\ast\Big(\frac{w_{BN}}{\sqrt{\sigma+\epsilon}}\text{Pad}(\textbf{x})+\text{Pad}(-\frac{w_{BN}\mu}{\sqrt{\sigma+\epsilon}}+b_{BN})\Big)\nonumber\\
&\qquad\qquad\qquad\qquad\qquad\qquad\qquad\qquad\quad\enspace+b_{L}\\
&\begin{aligned}
=\Big(w_{L}^\top\frac{w_{BN}}{\sqrt{\sigma+\epsilon}}\Big)&\ast\text{Pad}(\textbf{x})\\
+w_{L}&\ast\text{Pad}(-\frac{w_{BN}\mu}{\sqrt{\sigma+\epsilon}}+b_{BN})+b_{L}
\end{aligned}\\
&\begin{aligned}
=\underbrace{\Big(w_{L}^\top\frac{w_{BN}}{\sqrt{\sigma+\epsilon}}\Big)}_{w_\text{new}}&\ast\text{Pad}(\textbf{x})\\
&+\underbrace{\text{Conv}(\text{Pad}(-\frac{w_{BN}\mu}{\sqrt{\sigma+\epsilon}}+b_{BN}))}_{b_\text{new}}
\end{aligned}
\end{align}

In the equations above, $\ast$ stands for convolution. The new bias term is a full feature map, as opposed to a scalar as in linear layers without padding. The feature map does not depend on input \textbf{x} and is computed by putting a feature map (of the same size as \textbf{x}) with all features equal to $-\frac{w_{BN}\mu}{\sqrt{\sigma + \epsilon}}+b_{BN}$ through the original linear layer. Notice that if the padding value is nonzero, then the padding value of the canonized layer must be scaled by $\frac{\sqrt{\sigma + \epsilon}}{w_{BN}}$

We now give a simple example to help illustrate the problem and the proposed solution. Specifically, we set BN parameters $\mu=0,\sigma=1,\epsilon=0, w_{\text{BN}}=1, b_{\text{BN}}=1 $. Furthermore we define a single Convolutional filter with zero padding of width 1,
$w_{L}=\begin{bmatrix}
1 & 1\\
1 & 1\end{bmatrix}$ 
and no bias, $b_L=0$. Finally, we choose to show the case of a simple $3\times3$ input feature map

\begin{align}
\textbf{x} &=\begin{bmatrix}
1 & 2 & 3\\
2 & 3 & 4\\
3 & 4 & 5
\end{bmatrix}\\
Pad(BN(\textbf{x}))&=\begin{bmatrix}
0 & 0 & 0 & 0 & 0\\
0 & 2 & 3 & 4 & 0\\
0 & 3 & 4 & 5 & 0\\
0 & 4 & 5 & 6 & 0\\
0 & 0 & 0 & 0 & 0
\end{bmatrix}\\
w_{L} \ast Pad(BN(\textbf{x})) &= 
\begin{bmatrix}
2 & 5 & 7 & 4\\
5 & 12 & 16 & 9\\
7 & 16 & 20 & 11\\
4 & 9 & 11 & 6
\end{bmatrix}\\
=\begin{bmatrix}
1 & 3 & 5 & 3\\
3 & 8 & 12 & 7\\
5 & 12 & 16 & 9\\
3 & 7 & 9 & 5
\end{bmatrix}&+\begin{bmatrix}
1 & 2 & 2 & 1\\
2 & 4 & 4 & 2\\
2 & 4 & 4 & 2\\
1 & 2 & 2 & 1
\end{bmatrix}
\\
= w_{\text{new}} \ast Pad(\textbf{x}) &+ w_{L}\ast Pad(BNBias) + b_{L}
\end{align}

where and $BNBias=\begin{bmatrix}
1 &1 &1\\
1 &1 &1\\
1 &1 &1
\end{bmatrix}$ is the feature map composed of values equal to $-\frac{w_{BN}\mu}{\sqrt{\sigma + \epsilon}}+b_{BN}$


\section{Canonization of Relation Networks}
\label{sec:relation_network_details}
\paragraph{Architecture:} Relation Networks~\cite{santoro_simple_2017} are the state-of-the-art model architecture for the CLEVR dataset. It uses two separate encoders for image and text input. For the image, a simple convolutional neural network is used with 4 blocks, each containing a Convolutional layer, followed by a ReLU activation function and a BN layer. The text input is processed by a LSTM. The pixels from the feature map from the last Convolutional block from the image encoder are pair-wise concatenated along with their coordinates and the text encoding. This representation is then passed to a 4-layer fully connected network, summed up and then processed by a 3-layer fully connected network with ReLU activation.

\begin{figure*}[ht]
\centering
  \includegraphics[width=\linewidth]{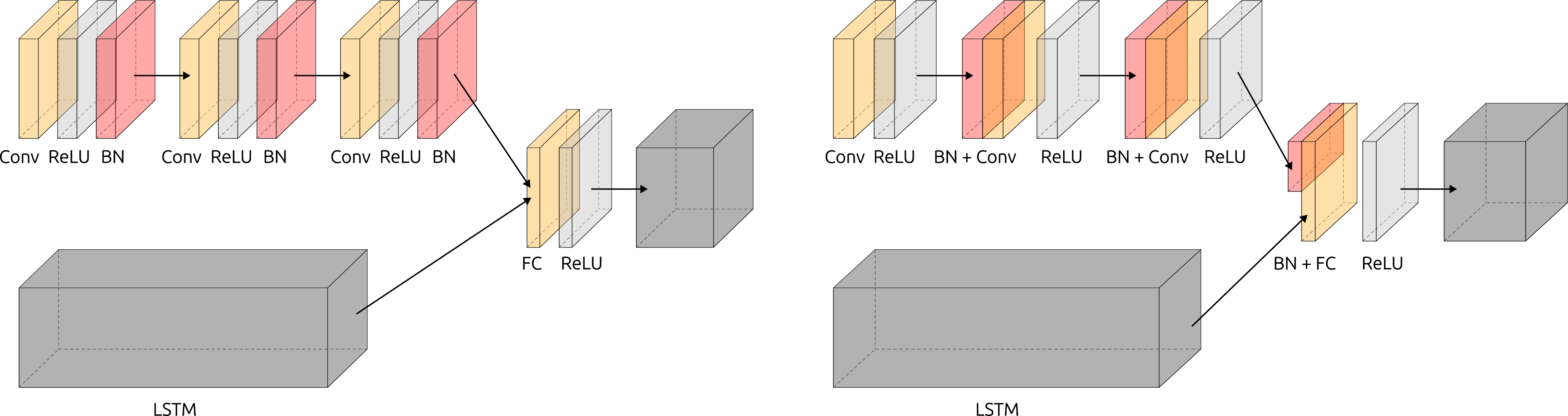}
\caption{Canonization of Relation Network. \emph{(Left):} Part from the original Relation Network. \emph{(Right):} Suggested canonization for the corresponding part of Relation Networks. BN layers are merged into the Convolutional layer at the beginning of the following block. The BN layer of the last block is merged into the following fully connected layer. However, only weights operating upon activations coming from the image encoder are updated, as outlined in Eq.~\eqref{eq:w_new_cat}.} 
\label{fig:rn_canonization}
\end{figure*}

\paragraph{Canonization:}Relation Networks, as implemented by the authors of~\cite{arras_clevr-xai_2022}, use \gls*{bn} layers at the end of each block \emph{not} directly after Convolutional layers. Therefore, we suggest to merge the \gls*{bn} layers with the Convolutional layers at the beginning of the following block. The \gls*{bn} layer of the last block of the image encoder can be merged into the following fully connected layer. However, attention as to be paid to make sure only weights operating on activations coming from the image encoder are updated, as outlined below. The proposed canonization of Relation Networks is visualized in Fig.~\ref{fig:rn_canonization}

\paragraph{Challenge:}In relation networks, the last BN layer of the image encoder has to be merged into a linear layer of the succeeding block, which takes as input a concatenation of image pairs, text and indices. Therefore, only the weights responsible for the activations coming from the image encoder have to be updated.

\paragraph{Incoming Activations:}
\begin{equation}
    \textbf{x} = \text{concat}
    \overbrace{\Big[
    \underbrace{[\textbf{x1}]}_{24}, 
    \underbrace{[\text{coord1}]}_{2}, 
    \underbrace{[\textbf{x2}]}_{24}, 
    \underbrace{[\text{coord2}]}_{2}, 
    \underbrace{[\text{question}]}_{128}
    \Big]}^{180}
\end{equation}
\newline
Only $\textbf{x1}$ and $\textbf{x2}$ pass the BN layer, i.e., indices $0:24$ and $26:50$ have to be updated. Here, the indexing $i:j$ signifies the elements with indices from $i$ to $j-1$, where the first element is indexed with $0$. In order to update only the relevant part of the weights of the linear layer $w_L$, we have to split them into:
\begin{equation}
    w_L = \text{concat}
    \Big[
    \underbrace{[w_L^{0:24}]}_{\textbf{x1}}, 
    \underbrace{[w_L^{24:26}]}_{\text{coord1}}, 
    \underbrace{[w_L^{26:50}]}_{\textbf{x2}}, 
    \underbrace{[w_L^{50:52}]}_{\text{coord2}}, 
    \underbrace{[w_L^{52:180}]}_{\text{text}}
    \Big]
\end{equation}
\newline
Using Eq.~\eqref{appeq:can_bn_lin_end}, each relevant weight part can than be updated as follows:

  \begin{equation}
    w_{L_\text{new}}^{0:24}=\frac{{w_L^{0:24}}^\top w_{BN}}{\sqrt{\sigma + \epsilon}}
  \end{equation}
  \begin{equation}
    w_{L_\text{new}}^{26:50}=\frac{{w_L^{26:50}}^\top w_{BN}}{\sqrt{\sigma + \epsilon}}
  \end{equation}
This gives a new weight matrix:
\begin{equation}
    w_{L_\text{new}} = \text{concat}
    \Big[
    \underbrace{[w_{L_\text{new}}^{0:24}]}_{\textbf{x1}}, 
    \underbrace{[w_L^{24:26}]}_{\text{coord1}}, 
    \underbrace{[w_{L_\text{new}}^{26:50}]}_{\textbf{x2}}, 
    \underbrace{[w_L^{50:52}]}_{\text{coord2}}, 
    \underbrace{[w_L^{52:180}]}_{\text{text}}
    \Big]
    \label{eq:w_new_cat}
\end{equation}
Similarly, the new bias can be calculated as:
\begin{equation}
    b_{L_\text{new}} = w_L^{0:24} b_\text{linBN} + w_L^{26:50} b_\text{linBN} + b_C
\end{equation}
 with $b_\text{linBN}=b_{BN} - \frac{w_BN \cdot \mu}{\sqrt{\sigma + \epsilon}}$

\section{Composites}
The composites, i.e., pre-defined layer-to-rule assignments as suggested in the literature, that we used in the paper, are described in Tab.~\ref{tab:composites}.
\label{ap:composites}

\begin{table}[ht]
    \centering

\begin{tabular}{l|l|l}
\toprule
Composite &            Layer Type &       Rule \\
\midrule
\multirow{2}{*}{LRP-$\epsilon+$ }         &         Convolutional &  $\alpha1\beta0$-rule \\
        &       Fully Connected &   $\epsilon$-rule \\
\midrule
\multirow{2}{*}{LRP-$\alpha2\beta1$ }       &         Convolutional &  $\alpha2\beta1$-rule \\
      &       Fully Connected &   $\epsilon$-rule \\
\midrule
\multirow{3}{*}{LRP-Custom (RN)} &   First Convolutional &   box-rule \\
 &  Other Convolutionals &  $\alpha1\beta0$-rule \\
 &       Fully Connected &  $\alpha1\beta0$-rule \\
\bottomrule
\end{tabular}
\caption{\label{tab:composites} Details for composites used in our experiments.}
\end{table}

\section{Pascal VOC 2012 Experiments}
\label{sec:experiments_pascal}
\subsection{Dataset Description}
Pascal Visual Object Classes (VOC) 2012 dataset has images from 20 categories, including 5717 training samples and 5823 validation samples with bounding box annotations, along with a private test set. From those, 1464 training samples and 1449 validation samples are annotated with binary segmentation masks. As opposed to ILSVRC2017, the images are much more diverse in composition. Many images contain multiple instances of several categories. The dataset does not suffer from the center-bias mentioned for ILSVRC2017. In the experiments, we use the validation samples with segmentation masks. Due to the robustness of models to input perturbations, the faithfulness correlation scores are very low, even entirely zero for some models. In order to obtain more meaningful results, we report faithfulness correlation scores with bigger perturbations compared to the ILSVRC2017 experiments.

\subsection{Models}  We evaluate explanations on VGG-16, ResNet-18, ResNet-50, EfficientNet-B0, EfficientNet-B4, DenseNet-121 and DenseNet-161. We fine tune models using the full training set, using pre-trained models from the PyTorch model zoo. We use stochastic gradient descent as the learning algorithm, with a cosine annealing learning rate scheduler~\cite{loshcilov_cosine-annealing_2016}. We use sum of binary cross entropy losses as the loss, and train the networks until convergence. We opted to use a smaller learning rate for pretrained parameters. 

\subsection{Results} 
The results are shown in Tables~\ref{tab:vgg16_VOC}~-~\ref{tab:densenet161_VOC}. The results suggest that canonization increases performance in the localization metrics and partially for faithfulness metrics. 
For the robustness metrics, canonization helps for Excitation Backpropagation. However, it makes robustness scores improve by a bigger margin for all methods for DenseNet models. 
Similar to the results for ILSVRC2017, DenseNet models seem to be affected negatively in their complexity scores when canonized. For other model architectures, complexity measures are also uniformly improved by canonization.

\begin{table*}[ht]
    \centering
    \resizebox{\textwidth}{!}{ \setlength{\tabcolsep}{5pt}
\begin{tabular}{l||cc|cccc|cccc|cc|cccc}
\toprule
{} & \multicolumn{2}{c|}{Complexity} & \multicolumn{4}{c|}{Faithfulness} & \multicolumn{4}{c|}{Localization} & \multicolumn{2}{c|}{Random.} & \multicolumn{4}{c}{Robustness} \\
{} & \multicolumn{2}{c|}{$\uparrow$\ Spars.} & \multicolumn{2}{c}{$\uparrow$ Corr.} & \multicolumn{2}{c|}{$\uparrow$ AoPC} & \multicolumn{2}{c}{$\uparrow$ RMA} & \multicolumn{2}{c|}{$\uparrow$ RRA} & \multicolumn{2}{c|}{$\downarrow$ Logit} & \multicolumn{2}{c}{$\downarrow$ Avg. Sens.} & \multicolumn{2}{c}{$\downarrow$ Max Sens.} \\
canonized &         no &   yes &           no &   yes &    no &   yes &           no &   yes &    no &   yes &      no &   yes &         no &   yes &        no &   yes \\
\midrule
EB  &  0.59  &  \textbf{0.60}  &  0.04  &  0.04  &  0.21  &  \textbf{0.22} &  0.33  &  \textbf{0.34}  &  0.40  &  \textbf{0.41}  &  1.00  &  1.00  &  0.24  &  \textbf{0.20}  &  0.26  &  \textbf{0.21}\\
LRP-$\alpha2\beta1$  &  0.75  &  \textbf{0.86}  &  \textbf{0.04}  &  0.03  &  0.27  &  0.27  &  0.38  &  \textbf{0.44}  &  0.40  &  \textbf{0.41}  &  \textbf{0.53}  &  0.64  &  0.53  &  0.53  &  0.80  &  \textbf{0.78}\\
LRP-$\varepsilon$+  &  0.59  &  \textbf{0.68}  &  0.05  &  0.05  & 0.25  &  \textbf{0.26}  &  0.36  &  \textbf{0.40}  &  0.46  &  \textbf{0.47}  &  0.49  &  0.49  &  0.51  &  0.51  &  0.80  &  \textbf{0.78}\\
\bottomrule
\end{tabular}

}
    \caption{\label{tab:vgg16_VOC}Results for Pascal VOC XAI evaluation with VGG-16. Arrows indicate whether high ($\uparrow$) or low ($\downarrow$) are better. Best results are shown in bold.}
\end{table*}

\begin{table*}[ht]
    \centering
    \resizebox{\textwidth}{!}{ \setlength{\tabcolsep}{5pt}
\begin{tabular}{l||cc|cccc|cccc|cc|cccc}
\toprule
{} & \multicolumn{2}{c|}{Complexity} & \multicolumn{4}{c|}{Faithfulness} & \multicolumn{4}{c|}{Localization} & \multicolumn{2}{c|}{Random.} & \multicolumn{4}{c}{Robustness} \\
{} & \multicolumn{2}{c|}{$\uparrow$\ Spars.} & \multicolumn{2}{c}{$\uparrow$ Corr.} & \multicolumn{2}{c|}{$\uparrow$ AoPC} & \multicolumn{2}{c}{$\uparrow$ RMA} & \multicolumn{2}{c|}{$\uparrow$ RRA} & \multicolumn{2}{c|}{$\downarrow$ Logit} & \multicolumn{2}{c}{$\downarrow$ Avg. Sens.} & \multicolumn{2}{c}{$\downarrow$ Max Sens.} \\
canonized &         no &   yes &           no &   yes &    no &   yes &           no &   yes &    no &   yes &      no &   yes &         no &   yes &        no &   yes \\
\midrule
EB  &  0.53  &  \textbf{0.57}  &  0.03  &  \textbf{0.04}  &  0.22  &  \textbf{0.23}  &  0.31  &  \textbf{0.33}  &  0.40  &  \textbf{0.41}  &  \textbf{0.96}  &  0.97  &  0.17  &  \textbf{0.15}  &  0.19  &  \textbf{0.16}\\
LRP-$\alpha2\beta1$  &  0.70  &  \textbf{0.77}  &  0.03  &  0.03  &  0.23  &  0.23  &  0.34  &  \textbf{0.39}  &  0.36  &  \textbf{0.39}  &  0.65  &  \textbf{0.64}  &  0.48  &  0.48  &  0.77  &  0.77\\
LRP-$\varepsilon$+  &  0.56  &  \textbf{0.63}  &  0.03  &  0.03  &  0.22  &  0.22  &  0.31  &  \textbf{0.34}  &  0.39  &  \textbf{0.41}  &  0.68  &  0.68  &  0.46  &  \textbf{0.45}  &  0.77  &  \textbf{0.76}\\
\bottomrule
\end{tabular}

}
    \caption{\label{tab:resnet18_VOC}Results for Pascal VOC XAI evaluation with ResNet-18. Arrows indicate whether high ($\uparrow$) or low ($\downarrow$) are better. Best results are shown in bold.}
\end{table*}

\begin{table*}[ht]
    \centering
    \resizebox{\textwidth}{!}{ \setlength{\tabcolsep}{5pt}
\begin{tabular}{l||cc|cccc|cccc|cc|cccc}
\toprule
{} & \multicolumn{2}{c|}{Complexity} & \multicolumn{4}{c|}{Faithfulness} & \multicolumn{4}{c|}{Localization} & \multicolumn{2}{c|}{Random.} & \multicolumn{4}{c}{Robustness} \\
{} & \multicolumn{2}{c|}{$\uparrow$\ Spars.} & \multicolumn{2}{c}{$\uparrow$ Corr.} & \multicolumn{2}{c|}{$\uparrow$ AoPC} & \multicolumn{2}{c}{$\uparrow$ RMA} & \multicolumn{2}{c|}{$\uparrow$ RRA} & \multicolumn{2}{c|}{$\downarrow$ Logit} & \multicolumn{2}{c}{$\downarrow$ Avg. Sens.} & \multicolumn{2}{c}{$\downarrow$ Max Sens.} \\
canonized &         no &   yes &           no &   yes &    no &   yes &           no &   yes &    no &   yes &      no &   yes &         no &   yes &        no &   yes \\
\midrule
EB  &  0.55  &  \textbf{0.63}  &  0.04  &  0.04  &  0.20  &  \textbf{0.23}  &  0.30  &  \textbf{0.35}  &  0.38  &  \textbf{0.42}  &  0.96  &  \textbf{0.93}  &  0.20  &  \textbf{0.17}  &  0.21  &  \textbf{0.18}\\
LRP-$\alpha2\beta1$  &  0.73  &  \textbf{0.81}  &  0.03  &  0.03  &  0.24  &  0.24  &  0.38  &  \textbf{0.41}  &  0.39  &  0.39  &  \textbf{0.60}  &  0.62  &  \textbf{0.48}  &  0.49  &  0.74  &  \textbf{0.73}\\
LRP-$\varepsilon$+  &  0.60  &  \textbf{0.69}  &  0.04  &  0.04  &  0.24  &  \textbf{0.25}  &  0.35  &  \textbf{0.39}  &  0.44  &  0.44  &  0.60  &  \textbf{0.57}  &  0.46  &  0.46  &  0.75  &  \textbf{0.74}\\
\bottomrule
\end{tabular}

}
    \caption{\label{tab:resnet50_VOC}Results for Pascal VOC XAI evaluation with ResNet-50. Arrows indicate whether high ($\uparrow$) or low ($\downarrow$) are better. Best results are shown in bold.}
\end{table*}

\begin{table*}[ht]
    \centering
    \resizebox{\textwidth}{!}{ \setlength{\tabcolsep}{5pt}
\begin{tabular}{l||cc|cccc|cccc|cc|cccc}
\toprule
{} & \multicolumn{2}{c|}{Complexity} & \multicolumn{4}{c|}{Faithfulness} & \multicolumn{4}{c|}{Localization} & \multicolumn{2}{c|}{Random.} & \multicolumn{4}{c}{Robustness} \\
{} & \multicolumn{2}{c|}{$\uparrow$\ Spars.} & \multicolumn{2}{c}{$\uparrow$ Corr.} & \multicolumn{2}{c|}{$\uparrow$ AoPC} & \multicolumn{2}{c}{$\uparrow$ RMA} & \multicolumn{2}{c|}{$\uparrow$ RRA} & \multicolumn{2}{c|}{$\downarrow$ Logit} & \multicolumn{2}{c}{$\downarrow$ Avg. Sens.} & \multicolumn{2}{c}{$\downarrow$ Max Sens.} \\
canonized &         no &   yes &           no &   yes &    no &   yes &           no &   yes &    no &   yes &      no &   yes &         no &   yes &        no &   yes \\
\midrule
EB  &  0.55  &  \textbf{0.69}  &  0.01  &  \textbf{0.02}  &  0.13  &  \textbf{0.18}  &  0.26  &  \textbf{0.37}  &  0.30  &  \textbf{0.41}  &  1.00  &  \textbf{0.99}  &  0.43  &  \textbf{0.39}  &  0.50  &  \textbf{0.43}\\
LRP-$\alpha2\beta1$  &  0.73  &  \textbf{0.76}  &  0.00  &  0.00  &  \textbf{0.18}  &  0.11  &  \textbf{0.37}  &  0.34  &  \textbf{0.39}  &  0.34  &  0.63  &  \textbf{0.51}  &  0.58  &  0.58  &  0.80  &  \textbf{0.79}\\
LRP-$\varepsilon$+  &  0.51  &  \textbf{0.72}  &  0.02  &  0.02  &  0.17  &  \textbf{0.19}  &  0.30  &  \textbf{0.39}  &  0.41  &  \textbf{0.43}  &  \textbf{0.59}  &  0.67  &  \textbf{0.46}  &  0.49  &  0.77  &  0.77\\
\bottomrule
\end{tabular}

}
    \caption{\label{tab:efficientnetb0_VOC}Results for Pascal VOC XAI evaluation with EfficientNet-B0. Arrows indicate whether high ($\uparrow$) or low ($\downarrow$) are better. Best results are shown in bold.}
 \end{table*}

\begin{table*}[ht]
    \centering
    \resizebox{\textwidth}{!}{ \setlength{\tabcolsep}{5pt}
\begin{tabular}{l||cc|cccc|cccc|cc|cccc}
\toprule
{} & \multicolumn{2}{c|}{Complexity} & \multicolumn{4}{c|}{Faithfulness} & \multicolumn{4}{c|}{Localization} & \multicolumn{2}{c|}{Random.} & \multicolumn{4}{c}{Robustness} \\
{} & \multicolumn{2}{c|}{$\uparrow$\ Spars.} & \multicolumn{2}{c}{$\uparrow$ Corr.} & \multicolumn{2}{c|}{$\uparrow$ AoPC} & \multicolumn{2}{c}{$\uparrow$ RMA} & \multicolumn{2}{c|}{$\uparrow$ RRA} & \multicolumn{2}{c|}{$\downarrow$ Logit} & \multicolumn{2}{c}{$\downarrow$ Avg. Sens.} & \multicolumn{2}{c}{$\downarrow$ Max Sens.} \\
canonized &         no &   yes &           no &   yes &    no &   yes &           no &   yes &    no &   yes &      no &   yes &         no &   yes &        no &   yes \\
\midrule
EB  &  0.68  &  \textbf{0.77}  &  0.0  &  0.0  &  \textbf{0.14}  &  0.13  &  0.33  &  \textbf{0.36}  &  0.37  &  0.37  &  \textbf{0.97}  &  1.00  &  \textbf{0.32}  &  0.36  &  \textbf{0.37}  &  0.43\\
LRP-$\alpha2\beta1$  &  0.71  &  \textbf{0.75}  &  0.0  &  0.0  &  \textbf{0.12}  &  0.08  &  0.27  &  \textbf{0.32}  &  0.31  &  \textbf{0.33}  &  0.46  &  \textbf{0.42}  &  0.62  &  \textbf{0.61}  &  0.93  &  \textbf{0.91}\\
LRP-$\varepsilon$+  &  0.54  &  \textbf{0.77}  &  0.0  &  0.0  &  0.15  &  \textbf{0.16}  &  0.28  &  \textbf{0.40}  &  0.36  &  \textbf{0.42}  &  \textbf{0.51}  &  0.57  &  0.51  &  0.51  &  0.88  &  \textbf{0.85}\\
\bottomrule
\end{tabular}

}
    \caption{\label{tab:efficientnetb4_VOC}Results for Pascal VOC XAI evaluation with EfficientNet-B4. Arrows indicate whether high ($\uparrow$) or low ($\downarrow$) are better. Best results are shown in bold.}
\end{table*}

\begin{table*}[ht]
    \centering
    \resizebox{\textwidth}{!}{ \setlength{\tabcolsep}{5pt}
\begin{tabular}{l||cc|cccc|cccc|cc|cccc}
\toprule
{} & \multicolumn{2}{c|}{Complexity} & \multicolumn{4}{c|}{Faithfulness} & \multicolumn{4}{c|}{Localization} & \multicolumn{2}{c|}{Random.} & \multicolumn{4}{c}{Robustness} \\
{} & \multicolumn{2}{c|}{$\uparrow$\ Spars.} & \multicolumn{2}{c}{$\uparrow$ Corr.} & \multicolumn{2}{c|}{$\uparrow$ AoPC} & \multicolumn{2}{c}{$\uparrow$ RMA} & \multicolumn{2}{c|}{$\uparrow$ RRA} & \multicolumn{2}{c|}{$\downarrow$ Logit} & \multicolumn{2}{c}{$\downarrow$ Avg. Sens.} & \multicolumn{2}{c}{$\downarrow$ Max Sens.} \\
canonized &         no &   yes &           no &   yes &    no &   yes &           no &   yes &    no &   yes &      no &   yes &         no &   yes &        no &   yes \\
\midrule
EB  &  \textbf{0.63}  &  0.59  &  0.02  &  \textbf{0.04}  &  0.11  &  \textbf{0.25}  &  0.25  &  \textbf{0.37}  &  0.28  &  \textbf{0.48}  &  0.98  &  \textbf{0.75}  &  0.60  &  \textbf{0.20}  &  1.07  &  \textbf{0.22}\\
LRP-$\varepsilon$+  &  \textbf{0.70}  &  0.63  &  0.02  &  \textbf{0.04}  &  0.18  &  \textbf{0.24}  &  0.34  &  \textbf{0.36}  &  0.39  &  \textbf{0.44}  &  \textbf{0.35}  &  0.44  &  0.63  &  \textbf{0.50}  &  1.08  &  \textbf{0.77}\\
LRP-$\alpha2\beta1$  &  \textbf{0.83}  &  0.73  &  0.01  &  \textbf{0.03}  &  0.16  &  \textbf{0.24}  &  0.31  &  \textbf{0.39}  &  0.32  &  \textbf{0.41}  &  0.38  &  \textbf{0.36}  &  0.64  &  \textbf{0.51}  &  1.11  &  \textbf{0.76}\\
\bottomrule
\end{tabular}

}
    \caption{\label{tab:densenet121_VOC}Results for Pascal VOC XAI evaluation with DenseNet-121. Arrows indicate whether high ($\uparrow$) or low ($\downarrow$) are better. Best results are shown in bold.}
\end{table*}

\begin{table*}[ht]
    \centering
    \resizebox{\textwidth}{!}{ \setlength{\tabcolsep}{5pt}
\begin{tabular}{l||cc|cccc|cccc|cc|cccc}
\toprule
{} & \multicolumn{2}{c|}{Complexity} & \multicolumn{4}{c|}{Faithfulness} & \multicolumn{4}{c|}{Localization} & \multicolumn{2}{c|}{Random.} & \multicolumn{4}{c}{Robustness} \\
{} & \multicolumn{2}{c|}{$\uparrow$\ Spars.} & \multicolumn{2}{c}{$\uparrow$ Corr.} & \multicolumn{2}{c|}{$\uparrow$ AoPC} & \multicolumn{2}{c}{$\uparrow$ RMA} & \multicolumn{2}{c|}{$\uparrow$ RRA} & \multicolumn{2}{c|}{$\downarrow$ Logit} & \multicolumn{2}{c}{$\downarrow$ Avg. Sens.} & \multicolumn{2}{c}{$\downarrow$ Max Sens.} \\
canonized &         no &   yes &           no &   yes &    no &   yes &           no &   yes &    no &   yes &      no &   yes &         no &   yes &        no &   yes \\
\midrule
EB  &  \textbf{0.80}  &  0.58  &  0.01  &  \textbf{0.05}  &  0.05  &  \textbf{0.23}  &  0.11  &  \textbf{0.35}  &  0.17  &  \textbf{0.47}  &  0.99  &  \textbf{0.77}  &  0.57  &  \textbf{0.19}  &  0.84  &  \textbf{0.21}\\
LRP-$\varepsilon$+  &  \textbf{0.67}  &  0.63  &  0.02  &  \textbf{0.05}  &  0.17  &  \textbf{0.22}  &  0.34  &  \textbf{0.36}  &  0.40  &  \textbf{0.44}  &  \textbf{0.32}  &  0.49  &  0.62  &  \textbf{0.48}  &  1.06  &  \textbf{0.74}\\
LRP-$\alpha2\beta1$  &  \textbf{0.82}  &  0.74  &  0.01  &  \textbf{0.03}  &  0.16  &  \textbf{0.21}  &  0.31  &  \textbf{0.39}  &  0.33  &  \textbf{0.41}  &  0.52  &  \textbf{0.45}  &  0.64  &  \textbf{0.50}  &  1.13  &  \textbf{0.74}\\
\bottomrule
\end{tabular}

}
    \caption{\label{tab:densenet161_VOC}Results for Pascal VOC XAI evaluation with DenseNet-161. Arrows indicate whether high ($\uparrow$) or low ($\downarrow$) are better. Best results are shown in bold.}
\end{table*}

\section{ILSVRC2017 Experiments}
\subsection{Classes}
\label{app:sec:imagenet_classes}
In our experiments we considered the following 50 classes that were picked randomly:

\emph{``Bernese\_mountain\_dog'',  ``Christmas\_stocking'',  ``Gila\_monster'',  ``Shetland\_sheepdog'',  ``Windsor\_tie'',  ``amphibian'',  ``ant'',  ``bubble'',  ``cassette'',  ``cicada'',  ``collie'',  ``crossword\_puzzle'',  ``dalmatian'',  ``eft'',  ``file'',  ``flute'',  ``goldfish'',  ``gorilla'',  ``gown'',  ``grasshopper'',  ``green\_snake'',  ``gyromitra'',  ``hammer'',  ``hen\_of\_the\_woods'',  ``indigo\_bunting'',  ``kimono'',  ``magnetic\_compass'',  ``mongoose'',  ``mountain\_tent'',  ``otterhound'',  ``palace'',  ``patio'',  ``pencil\_sharpener'',  ``platypus'',  ``pomegranate'',  ``pool\_table'',  ``redshank'',  ``refrigerator'',  ``rhinoceros\_beetle'',  ``screw'',  ``screwdriver'',  ``shoe\_shop'',  ``shopping\_basket'',  ``stage'',  ``standard\_poodle'',  ``stethoscope'',  ``toaster'',  ``tree\_frog'',  ``vase'',  ``wolf\_spider''}.

\subsection{Additional Results}
In Tables~\ref{tab:vgg16}~-~\ref{tab:densenet161} we show additional results for our experiments with the ILSVRC2017 dataset. 
Specifically, in addition to the architectures evaluated in the main paper, we present results for ResNet50, EfficientNet-B4 and DenseNet-161.
Moreover, we include results for Faithfulness Correlation~\cite{bhatt_evaluating_2020} and Max Sensitivity~\cite{yeh__2019}.

\begin{table*}[ht]
    \centering
    \resizebox{\textwidth}{!}{
\setlength{\tabcolsep}{5pt}
\begin{tabular}{l||cc|cccc|cccc|cc|cccc}
\toprule
{} & \multicolumn{2}{c|}{Complexity} & \multicolumn{4}{c|}{Faithfulness} & \multicolumn{4}{c|}{Localization} & \multicolumn{2}{c|}{Random.} & \multicolumn{4}{c}{Robustness} \\
{} & \multicolumn{2}{c|}{$\uparrow$\ Spars.} & \multicolumn{2}{c}{$\uparrow$ Corr.} & \multicolumn{2}{c|}{$\uparrow$ AoPC} & \multicolumn{2}{c}{$\uparrow$ RMA} & \multicolumn{2}{c|}{$\uparrow$ RRA} & \multicolumn{2}{c|}{$\downarrow$ Logit} & \multicolumn{2}{c}{$\downarrow$ Avg. Sens.} & \multicolumn{2}{c}{$\downarrow$ Max Sens.} \\
canonized &         no &   yes &           no &   yes &    no &   yes &           no &   yes &    no &   yes &      no &   yes &         no &   yes &        no &   yes \\
\midrule

EB                  &       0.57 &  \textbf{0.59} &         0.06 &  0.06 &  0.35 &  \textbf{0.36} &         0.68 &  \textbf{0.70} &  0.70 &  \textbf{0.71} &    1.00 &  1.00 &       0.22 &  \textbf{0.18} &      0.23 &  \textbf{0.20} \\
LRP-$\alpha2\beta1$ &       0.70 &  \textbf{0.84} &         \textbf{0.05} &  0.03 &  0.38 &  \textbf{0.39} &         0.65 &  \textbf{0.77} &  0.63 &  \textbf{0.67} &    \textbf{0.59} &  0.66 &       \textbf{0.31} &  0.34 &      \textbf{0.34} &  0.37 \\
LRP-$\varepsilon$+     &       0.51 &  \textbf{0.62} &         \textbf{0.09} &  0.08 &  0.36 &  \textbf{0.39} &         0.64 &  \textbf{0.71} &  0.69 &  \textbf{0.71} &    0.57 &  \textbf{0.54} &       \textbf{0.19} &  0.21 &      \textbf{0.21} &  0.24 \\

\bottomrule
\end{tabular}
}
    \caption{\label{tab:vgg16} Results for ILSVRC2017 XAI evaluation with VGG-16. Arrows indicate whether high ($\uparrow$) or low ($\downarrow$) are better. Best results are shown in bold.}
\end{table*}

\begin{table*}[ht]
    \centering
    \resizebox{\textwidth}{!}{ 
\setlength{\tabcolsep}{5pt}
\begin{tabular}{l||cc|cccc|cccc|cc|cccc}
\toprule
{} & \multicolumn{2}{c|}{Complexity} & \multicolumn{4}{c|}{Faithfulness} & \multicolumn{4}{c|}{Localization} & \multicolumn{2}{c|}{Random.} & \multicolumn{4}{c}{Robustness} \\
{} & \multicolumn{2}{c|}{$\uparrow$\ Spars.} & \multicolumn{2}{c}{$\uparrow$ Corr.} & \multicolumn{2}{c|}{$\uparrow$ AoPC} & \multicolumn{2}{c}{$\uparrow$ RMA} & \multicolumn{2}{c|}{$\uparrow$ RRA} & \multicolumn{2}{c|}{$\downarrow$ Logit} & \multicolumn{2}{c}{$\downarrow$ Avg. Sens.} & \multicolumn{2}{c}{$\downarrow$ Max Sens.} \\
canonized &         no &   yes &           no &   yes &    no &   yes &           no &   yes &    no &   yes &      no &   yes &         no &   yes &        no &   yes \\
\midrule
EB                  &       0.55 &  \textbf{0.57} &         0.03 &  \textbf{0.04} &  0.29 &  0.29 &         0.66 &  \textbf{0.67} &  0.68 &  \textbf{0.69} &    0.97 &  0.97 &       0.16 &  \textbf{0.14} &      0.18 &  \textbf{0.15} \\
LRP-$\alpha2\beta1$ &       0.67 &  \textbf{0.76} &         \textbf{0.04} &  0.03 &  0.32 &  0.32 &         0.69 &  \textbf{0.75} &  0.65 &  \textbf{0.67} &    \textbf{0.65} &  0.61 &       \textbf{0.21} &  0.26 &      \textbf{0.22} &  0.28 \\
LRP-$\varepsilon$+     &       0.51 &  \textbf{0.58} &         0.04 &  0.04 &  0.30 &  0.30 &         0.65 &  \textbf{0.69} &  0.69 &  \textbf{0.70} &    0.70 &  0.70 &       \textbf{0.14} &  0.15 &      \textbf{0.15} &  0.16 \\
\bottomrule
\end{tabular}
}
    \caption{\label{tab:resnet18}Results for ILSVRC2017 XAI evaluation with ResNet-18. Arrows indicate whether high ($\uparrow$) or low ($\downarrow$) are better. Best results are shown in bold.}
\end{table*}

\begin{table*}[ht]
    \centering
    \resizebox{\textwidth}{!}{
\setlength{\tabcolsep}{5pt}
\begin{tabular}{l||cc|cccc|cccc|cc|cccc}
\toprule
{} & \multicolumn{2}{c|}{Complexity} & \multicolumn{4}{c|}{Faithfulness} & \multicolumn{4}{c|}{Localization} & \multicolumn{2}{c|}{Random.} & \multicolumn{4}{c}{Robustness} \\
{} & \multicolumn{2}{c|}{$\uparrow$\ Spars.} & \multicolumn{2}{c}{$\uparrow$ Corr.} & \multicolumn{2}{c|}{$\uparrow$ AoPC} & \multicolumn{2}{c}{$\uparrow$ RMA} & \multicolumn{2}{c|}{$\uparrow$ RRA} & \multicolumn{2}{c|}{$\downarrow$ Logit} & \multicolumn{2}{c}{$\downarrow$ Avg. Sens.} & \multicolumn{2}{c}{$\downarrow$ Max Sens.} \\
canonized &         no &   yes &           no &   yes &    no &   yes &           no &   yes &    no &   yes &      no &   yes &         no &   yes &        no &   yes \\
\midrule
EB                  &       \textbf{0.72} &  0.64 &         0.02 &  \textbf{0.04} &  0.24 &  \textbf{0.36} &         0.65 &  \textbf{0.71} &  0.66 &  \textbf{0.69} &    0.95 &  \textbf{0.93} &       0.36 &  \textbf{0.17} &      0.42 &  \textbf{0.18} \\
LRP-$\alpha2\beta1$ &       0.71 &  \textbf{0.81} &         \textbf{0.04} &  0.01 &  0.37 &  0.37 &         0.72 &  \textbf{0.77} &  0.66 &  \textbf{0.67} &    \textbf{0.59} &  0.61 &       \textbf{0.25} &  0.30 &      \textbf{0.27} &  0.33 \\
LRP-$\varepsilon$+     &       0.57 &  \textbf{0.67} &         \textbf{0.05} &  0.04 &  0.37 &  0.37 &         0.70 &  \textbf{0.74} &  \textbf{0.72} &  0.71 &    0.61 &  \textbf{0.60} &       \textbf{0.15} &  0.18 &      \textbf{0.16} &  0.19 \\
\bottomrule
\end{tabular}
}
    \caption{\label{tab:resnet50}Results for ILSVRC2017 XAI evaluation with ResNet-50. Arrows indicate whether high ($\uparrow$) or low ($\downarrow$) are better. Best results are shown in bold.}
\end{table*}

\begin{table*}[ht]
    \centering
    \resizebox{\textwidth}{!}{ 
\setlength{\tabcolsep}{5pt}
\begin{tabular}{l||cc|cccc|cccc|cc|cccc}
\toprule
{} & \multicolumn{2}{c|}{Complexity} & \multicolumn{4}{c|}{Faithfulness} & \multicolumn{4}{c|}{Localization} & \multicolumn{2}{c|}{Random.} & \multicolumn{4}{c}{Robustness} \\
{} & \multicolumn{2}{c|}{$\uparrow$\ Spars.} & \multicolumn{2}{c}{$\uparrow$ Corr.} & \multicolumn{2}{c|}{$\uparrow$ AoPC} & \multicolumn{2}{c}{$\uparrow$ RMA} & \multicolumn{2}{c|}{$\uparrow$ RRA} & \multicolumn{2}{c|}{$\downarrow$ Logit} & \multicolumn{2}{c}{$\downarrow$ Avg. Sens.} & \multicolumn{2}{c}{$\downarrow$ Max Sens.} \\
canonized &         no &   yes &           no &   yes &    no &   yes &           no &   yes &    no &   yes &      no &   yes &         no &   yes &        no &   yes \\
\midrule
EB                  &       \textbf{0.85} &  0.70 &         0.00 &  0.0 &  0.24 &  \textbf{0.27} &         \textbf{0.79} &  0.72 &  \textbf{0.73} &  0.67 &    \textbf{0.99} &  1.00 &       0.42 &  \textbf{0.33} &      0.48 &  \textbf{0.37} \\
LRP-$\alpha2\beta1$ &       0.75 &  \textbf{0.77} &         0.00 &  0.0 &  \textbf{0.29} &  0.20 &         \textbf{0.79} &  0.73 &  \textbf{0.72} &  0.65 &    0.57 &  \textbf{0.51} &       \textbf{0.48} &  0.49 &      \textbf{0.52} &  0.54 \\
LRP-$\varepsilon$+     &       0.50 &  \textbf{0.73} &         \textbf{0.01} &  0.0 &  0.28 &  \textbf{0.30} &         0.69 &  \textbf{0.79} &  0.75 &  0.75 &    \textbf{0.61} &  0.65 &       \textbf{0.12} &  0.21 &      \textbf{0.13} &  0.23 \\

\bottomrule
\end{tabular}
}
    \caption{\label{tab:efficientnetb0}Results for ILSVRC2017 XAI evaluation with EfficientNet-B0. Arrows indicate whether high ($\uparrow$) or low ($\downarrow$) are better. Best results are shown in bold. We analyzed the low faithfulness correlation scores and found that the model was very robust towards input perturbation, with output values remaining unaffected.}
\end{table*}

\begin{table*}[ht]
    \centering
    \resizebox{\textwidth}{!}{ 
\setlength{\tabcolsep}{5pt}
\begin{tabular}{l||cc|cccc|cccc|cc|cccc}
\toprule
{} & \multicolumn{2}{c|}{Complexity} & \multicolumn{4}{c|}{Faithfulness} & \multicolumn{4}{c|}{Localization} & \multicolumn{2}{c|}{Random.} & \multicolumn{4}{c}{Robustness} \\
{} & \multicolumn{2}{c|}{$\uparrow$\ Spars.} & \multicolumn{2}{c}{$\uparrow$ Corr.} & \multicolumn{2}{c|}{$\uparrow$ AoPC} & \multicolumn{2}{c}{$\uparrow$ RMA} & \multicolumn{2}{c|}{$\uparrow$ RRA} & \multicolumn{2}{c|}{$\downarrow$ Logit} & \multicolumn{2}{c}{$\downarrow$ Avg. Sens.} & \multicolumn{2}{c}{$\downarrow$ Max Sens.} \\
canonized &         no &   yes &           no &   yes &    no &   yes &           no &   yes &    no &   yes &      no &   yes &         no &   yes &        no &   yes \\
\midrule
EB                      &       \textbf{0.84} &  0.77 &          0.0 &  0.0 &  \textbf{0.20} &  0.19 &         0.64 &  \textbf{0.69} &  \textbf{0.68} &  0.66 &    \textbf{0.90} &  1.0 &       \textbf{0.30} &  0.35 &      \textbf{0.33} &  0.40 \\
LRP-$\alpha2\beta1$     &       0.77 &  \textbf{0.79} &          0.0 &  0.0 &  \textbf{0.15} &  0.13 &         0.53 &  \textbf{0.67} &  0.61 &  \textbf{0.64} &    0.43 &  \textbf{0.4}&       0.61 &  \textbf{0.53} &      0.68 &  \textbf{0.59} \\
LRP-$\varepsilon+$      &       0.56 &  \textbf{0.77} &          0.0 &  0.0 &  0.13 &  \textbf{0.24} &         0.56 &  \textbf{0.76} &  0.62 &  \textbf{0.70} &    0.54 &  \textbf{0.5}&       \textbf{0.14} &  0.23 &      \textbf{0.15} &  0.26 \\
\bottomrule
\end{tabular}
}
    \caption{\label{tab:efficientnetb4}Results for ILSVRC2017 XAI evaluation with EfficientNet-B4. Arrows indicate whether high ($\uparrow$) or low ($\downarrow$) are better. Best results are shown in bold. We analyzed the low faithfulness correlation scores and found that the model was very robust towards input perturbation, with output values remaining unaffected.}
\end{table*}

\begin{table*}[ht]
    \centering
    \resizebox{\textwidth}{!}{ \setlength{\tabcolsep}{5pt}
\begin{tabular}{l||cc|cccc|cccc|cc|cccc}
\toprule
{} & \multicolumn{2}{c|}{Complexity} & \multicolumn{4}{c|}{Faithfulness} & \multicolumn{4}{c|}{Localization} & \multicolumn{2}{c|}{Random.} & \multicolumn{4}{c}{Robustness} \\
{} & \multicolumn{2}{c|}{$\uparrow$\ Spars.} & \multicolumn{2}{c}{$\uparrow$ Corr.} & \multicolumn{2}{c|}{$\uparrow$ AoPC} & \multicolumn{2}{c}{$\uparrow$ RMA} & \multicolumn{2}{c|}{$\uparrow$ RRA} & \multicolumn{2}{c|}{$\downarrow$ Logit} & \multicolumn{2}{c}{$\downarrow$ Avg. Sens.} & \multicolumn{2}{c}{$\downarrow$ Max Sens.} \\
canonized &         no &   yes &           no &   yes &    no &   yes &           no &   yes &    no &   yes &      no &   yes &         no &   yes &        no &   yes \\
\midrule
EB                  &       \textbf{0.66} &  0.62 &         0.01 &  \textbf{0.03} &  0.15 &  \textbf{0.31} &         0.53 &  \textbf{0.73} &  0.58 &  \textbf{0.72} &    0.75 &  \textbf{0.89} &       0.57 &  \textbf{0.17} &      1.05 &  \textbf{0.19} \\
LRP-$\alpha2\beta1$ &       \textbf{0.82} &  0.81 &         0.01 &  \textbf{0.02} &  0.25 &  \textbf{0.33} &         0.68 &  \textbf{0.81} &  0.64 &  \textbf{0.71} &    0.40 &  \textbf{0.44} &       0.65 &  \textbf{0.28} &      1.30 &  \textbf{0.31} \\
LRP-$\varepsilon$+     &       \textbf{0.67} &  0.66 &         0.01 &  \textbf{0.03} &  0.26 &  \textbf{0.33} &         0.71 &  \textbf{0.77} &  0.70 &  \textbf{0.74} &    0.39 &  \textbf{0.48} &       0.63 &  \textbf{0.19} &      1.23 &  \textbf{0.21} \\
\bottomrule
\end{tabular}}
    \caption{\label{tab:densenet121}Results for ILSVRC2017 XAI evaluation with DenseNet-121. Arrows indicate whether high ($\uparrow$) or low ($\downarrow$) are better. Best results are shown in bold.}
\end{table*}

\begin{table*}[ht]
    \centering
    \resizebox{\textwidth}{!}{ 
\setlength{\tabcolsep}{5pt}
\begin{tabular}{l||cc|cccc|cccc|cc|cccc}
\toprule
{} & \multicolumn{2}{c|}{Complexity} & \multicolumn{4}{c|}{Faithfulness} & \multicolumn{4}{c|}{Localization} & \multicolumn{2}{c|}{Random.} & \multicolumn{4}{c}{Robustness} \\
{} & \multicolumn{2}{c|}{$\uparrow$\ Spars.} & \multicolumn{2}{c}{$\uparrow$ Corr.} & \multicolumn{2}{c|}{$\uparrow$ AoPC} & \multicolumn{2}{c}{$\uparrow$ RMA} & \multicolumn{2}{c|}{$\uparrow$ RRA} & \multicolumn{2}{c|}{$\downarrow$ Logit} & \multicolumn{2}{c}{$\downarrow$ Avg. Sens.} & \multicolumn{2}{c}{$\downarrow$ Max Sens.} \\
canonized &         no &   yes &           no &   yes &    no &   yes &           no &   yes &    no &   yes &      no &   yes &         no &   yes &        no &   yes \\
\midrule
EB                  &       \textbf{0.86} &  0.61 &         0.00 &  \textbf{0.03} &  0.05 &  \textbf{0.30} &         0.25 &  \textbf{0.71} &  0.46 &  \textbf{0.71} &    \textbf{0.86} &  0.90 &       0.56 &  \textbf{0.17} &      0.80 &  \textbf{0.18} \\
LRP-$\alpha2\beta1$ &       0.81 &  \textbf{0.82} &         0.01 &  0.01 &  0.25 &  \textbf{0.32} &         0.67 &  \textbf{0.82} &  0.65 &  \textbf{0.71} &    \textbf{0.34} &  0.45 &       0.65 &  \textbf{0.29} &      1.29 &  \textbf{0.33} \\
LRP-$\varepsilon$+     &       0.64 &  \textbf{0.66} &         0.02 &  \textbf{0.03} &  0.25 &  \textbf{0.32} &         0.70 &  \textbf{0.76} &  0.70 &  \textbf{0.74} &    \textbf{0.36} &  0.47 &       0.62 &  \textbf{0.18} &      1.19 &  \textbf{0.19} \\
\bottomrule
\end{tabular}}
    \caption{\label{tab:densenet161}Results for ILSVRC2017 XAI evaluation with DenseNet-161. Arrows indicate whether high ($\uparrow$) or low ($\downarrow$) are better. Best results are shown in bold.}
\end{table*}



\section{Additional CLEVR-XAI Results}
In Table~\ref{tab:clevr_max_norm}, we show additional results for our CLEVR-XAI experiments.
Specifically, in additional to \textit{pos-l2-norm-sq} pooling, we also present results for \textit{max-norm} pooling.


\begin{table*}[ht]
    \centering
    \resizebox{\textwidth}{!}{ \begin{tabular}{l||l|rr|rr|rr|rr|rr|rr}
\toprule
{}& {} & \multicolumn{2}{l|}{$\uparrow$ Complexity} & \multicolumn{2}{l|}{$\uparrow$ Faithfulness} & \multicolumn{2}{l|}{$\uparrow$ Local. (RRA)} & \multicolumn{2}{l|}{$\uparrow$ Local. (RMA)} & \multicolumn{2}{l|}{$\downarrow$ Robustness} & \multicolumn{2}{l}{$\downarrow$ Random.} \\
Questions & canonized &         no &   yes &           no &   yes &           no &   yes &           no &   yes &          no &   yes &            no &   yes \\
\midrule
\multirow{2}{*}{Simple} 
& EB            &       \textbf{0.92} &  0.79 &         0.50 &  0.50 &         \textbf{0.66} &  0.63 &         \textbf{0.56} &  0.38 &        1.34 &  \textbf{1.29} &    1.00 &  1.00 \\
& LRP-Custom    &       0.69 &  \textbf{0.82} &         0.52 &  0.52 &         0.71 &  0.71 &         0.34 &  \textbf{0.46} &        \textbf{1.19} &  1.24 &    0.99 &  0.99 \\
\midrule
\multirow{2}{*}{Complex} 
& EB            &       \textbf{0.91} &  0.81 &         \textbf{0.45} &  0.44 &         \textbf{0.67} &  0.64 &         \textbf{0.74} &  0.60 &        1.31 &  \textbf{1.22} &    0.99 &  0.99 \\
& LRP-Custom    &       0.70 &  \textbf{0.82} &         0.45 &  0.45 &         0.55 &  \textbf{0.64} &         0.48 &  \textbf{0.63} &        \textbf{1.16} &  1.20 &    \textbf{0.98} &  0.99 \\

\bottomrule
\end{tabular}
}
    \caption{\label{tab:clevr_max_norm}Results for CLEVR-XAI with Relation Network using \textit{max-norm} pooling. Arrows indicate whether high ($\uparrow$) or low ($\downarrow$) are better. Best results are shown in bold.}
\end{table*}


\section{Attribution Heatmaps}
\begin{figure*}[ht]
    \centering
  \includegraphics[width=.8\linewidth]{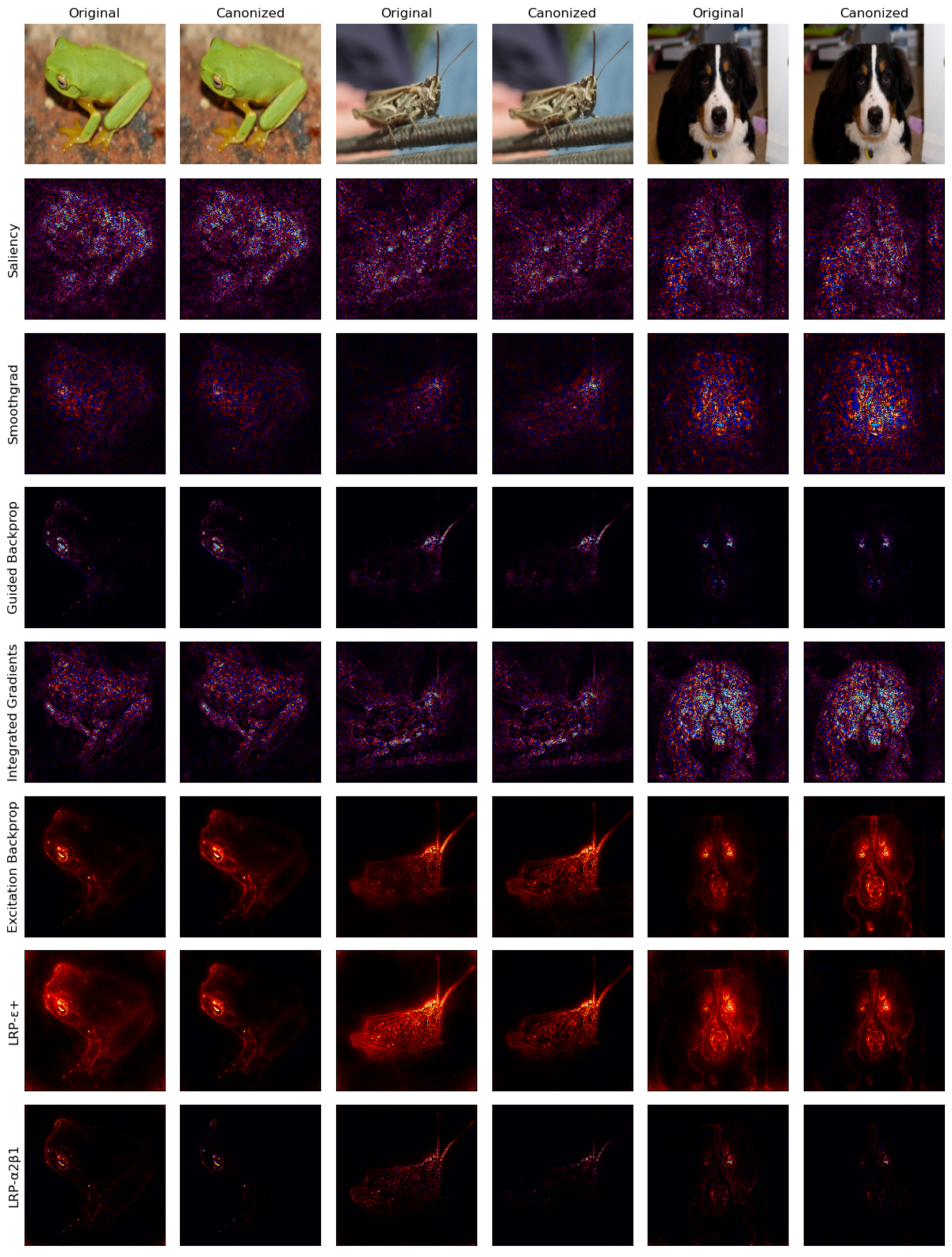}
  \caption{Attribution heatmaps with different XAI methods for VGG-16 model on ILSVRC2017 dataset.}
  \label{fig:attr_vgg16}
\end{figure*}

\begin{figure*}
    \centering
  \includegraphics[width=.8\linewidth]{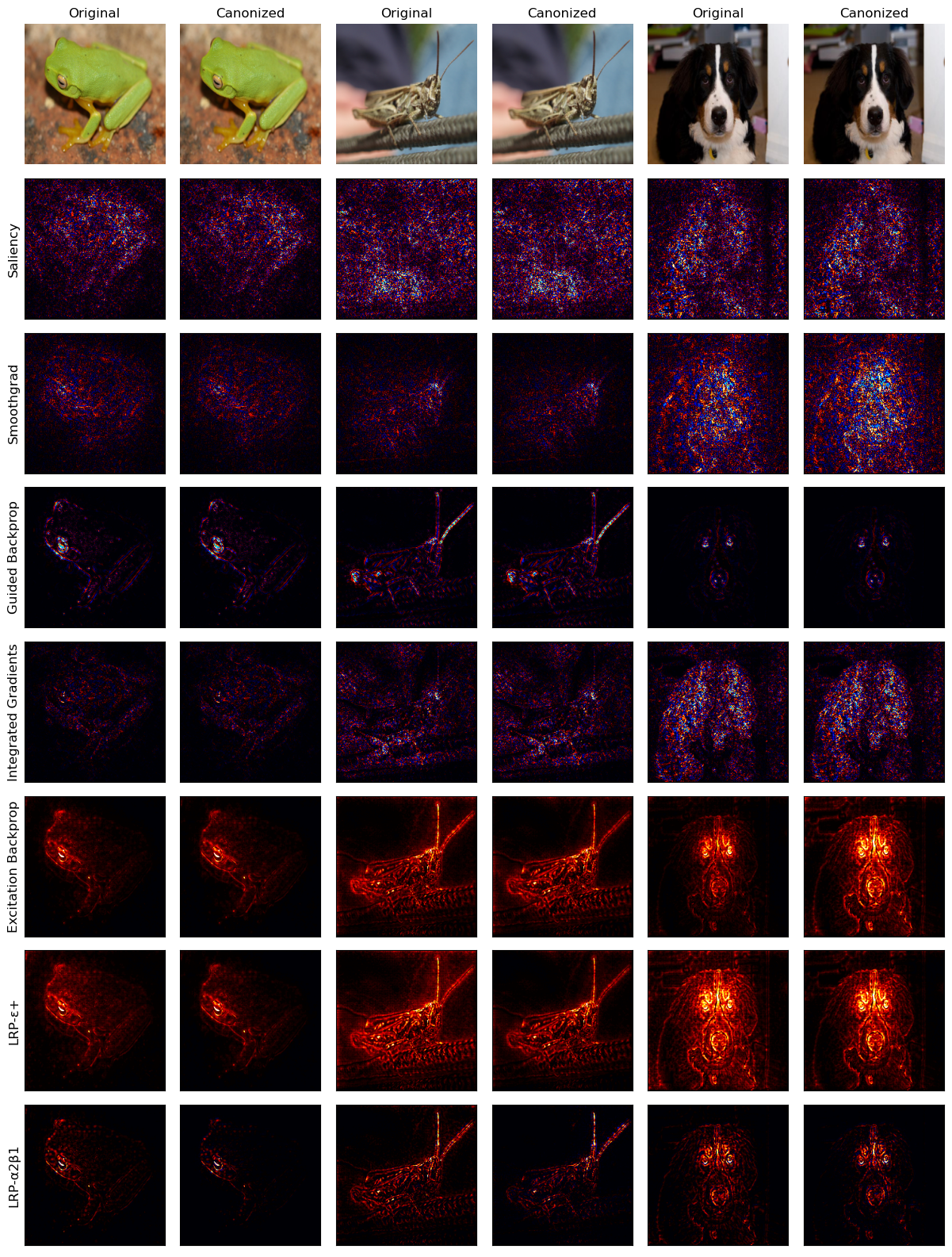}
  \caption{Attribution heatmaps with different XAI methods for ResNet-18 model on ILSVRC2017 dataset.}
  \label{fig:attr_resnet18}
\end{figure*}

\begin{figure*}[ht]
    \centering
  \includegraphics[width=.8\linewidth]{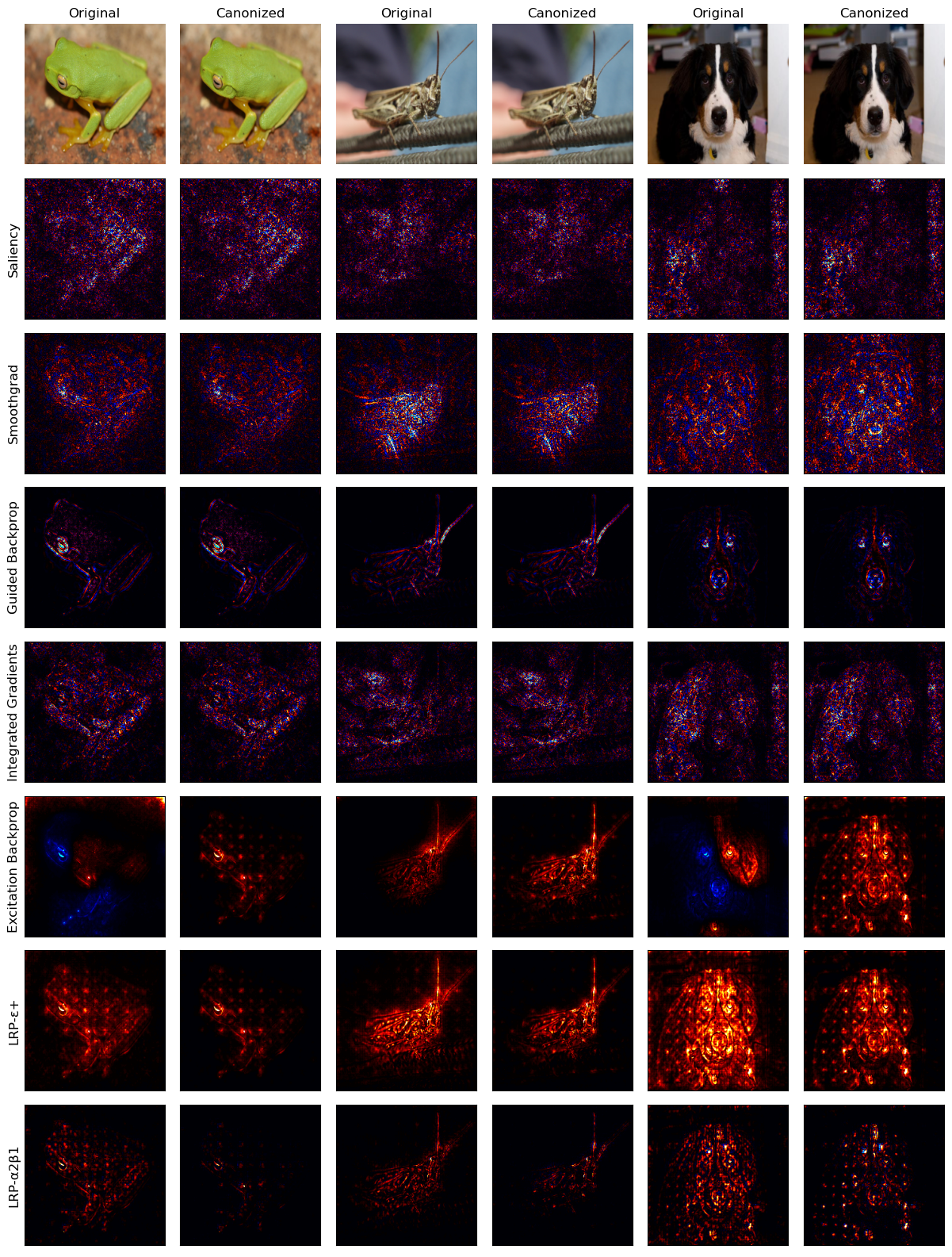}
  \caption{Attribution heatmaps with different XAI methods for ResNet-50 model on ILSVRC2017 dataset. The checkerboard pattern is due to the downsampling shortcuts in the network. We refer the reader to ~\cite{weber_beyond-explanations_2022} for details.}
  \label{fig:attr_resnet50}
\end{figure*}

\begin{figure*}[ht]
    \centering
  \includegraphics[width=.8\linewidth]{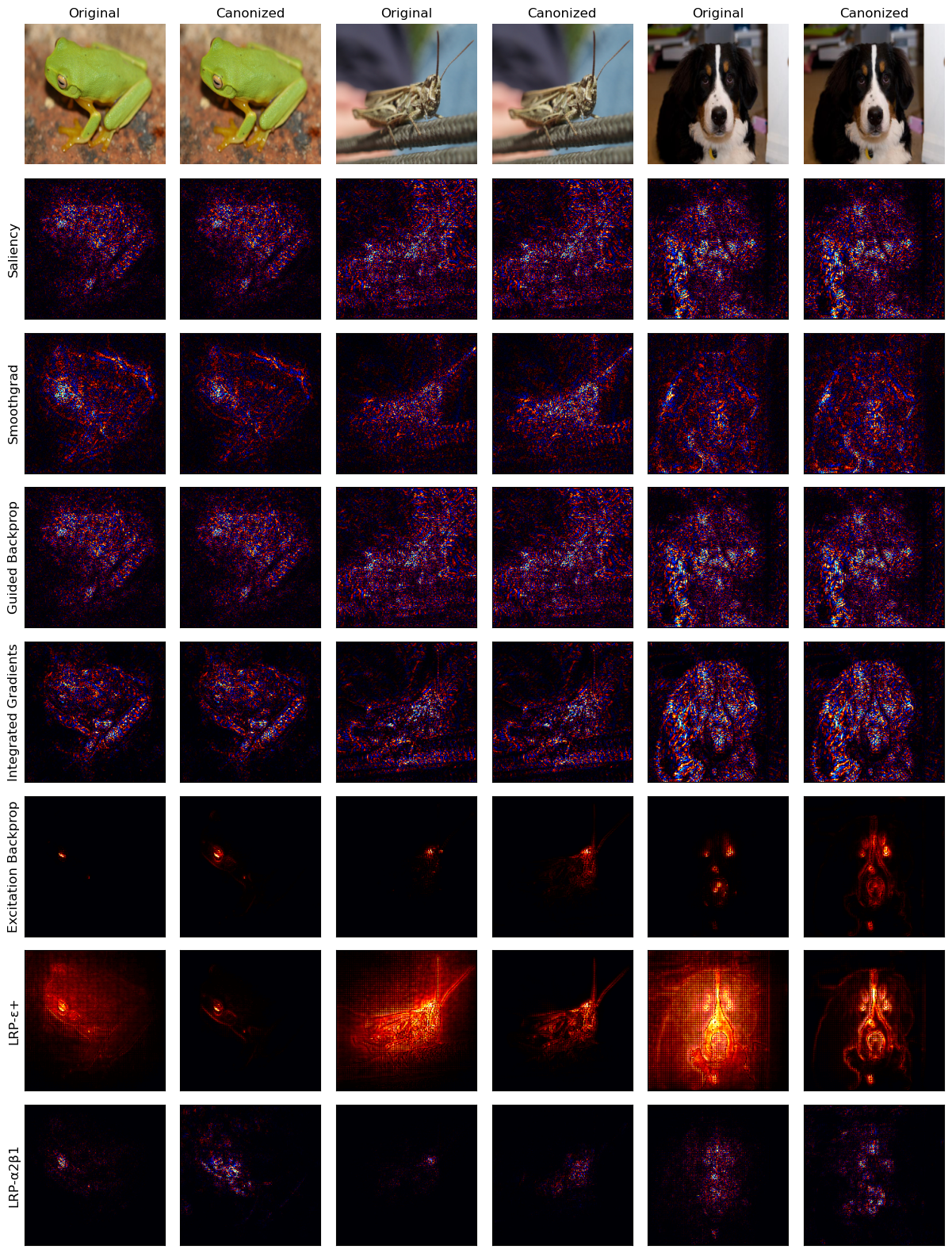}
  \caption{Attribution heatmaps with different XAI methods for EfficientNet-B0 model on ILSVRC2017 dataset.}
  \label{fig:attr_efficientnet_b0}
\end{figure*}

\begin{figure*}[ht]
    \centering
  \includegraphics[width=.8\linewidth]{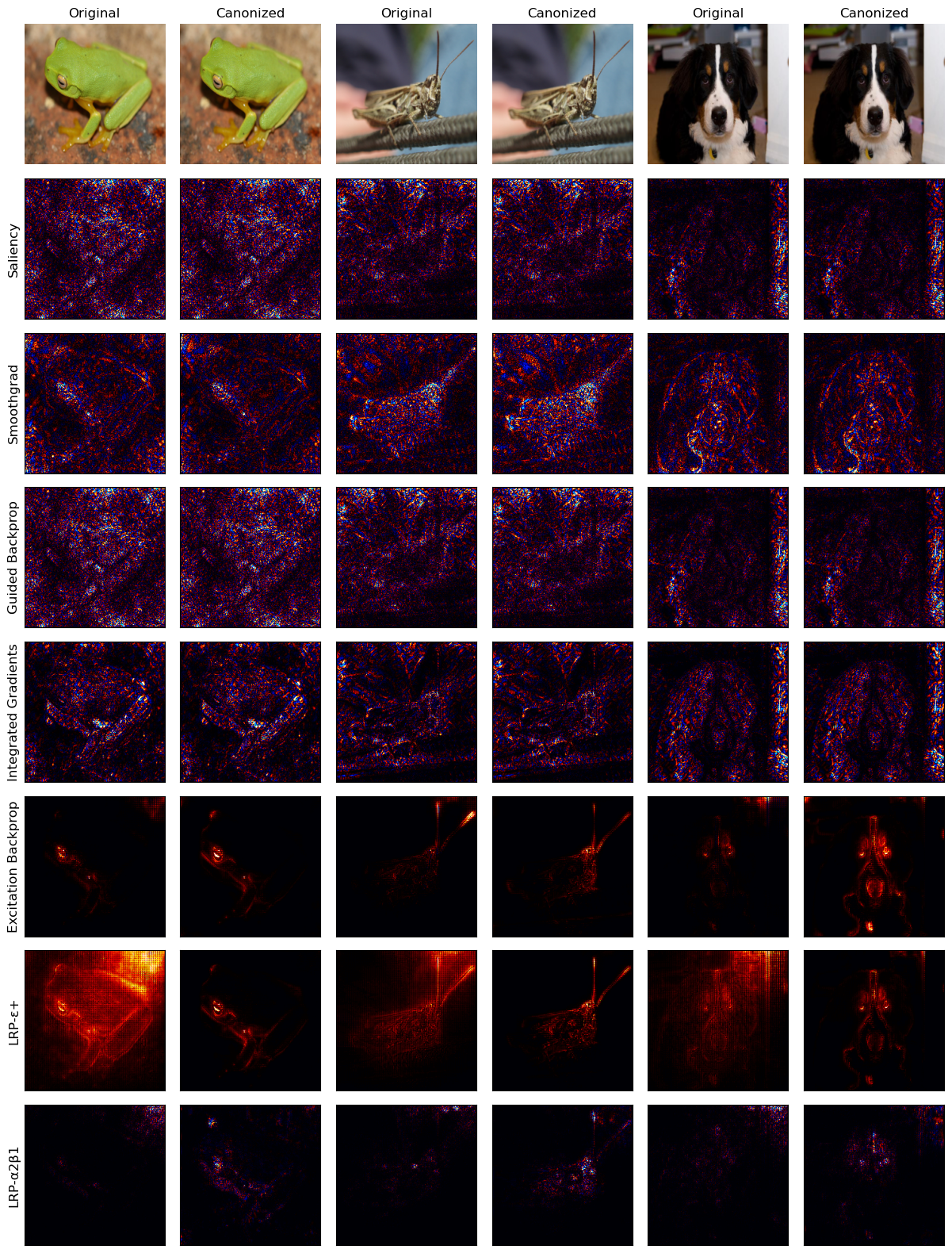}
  \caption{Attribution heatmaps with different XAI methods for EfficientNet-B4 model on ILSVRC2017 dataset.}
  \label{fig:attr_efficientnet_b4}
\end{figure*}

\begin{figure*}[ht]
    \centering
  \includegraphics[width=.8\linewidth]{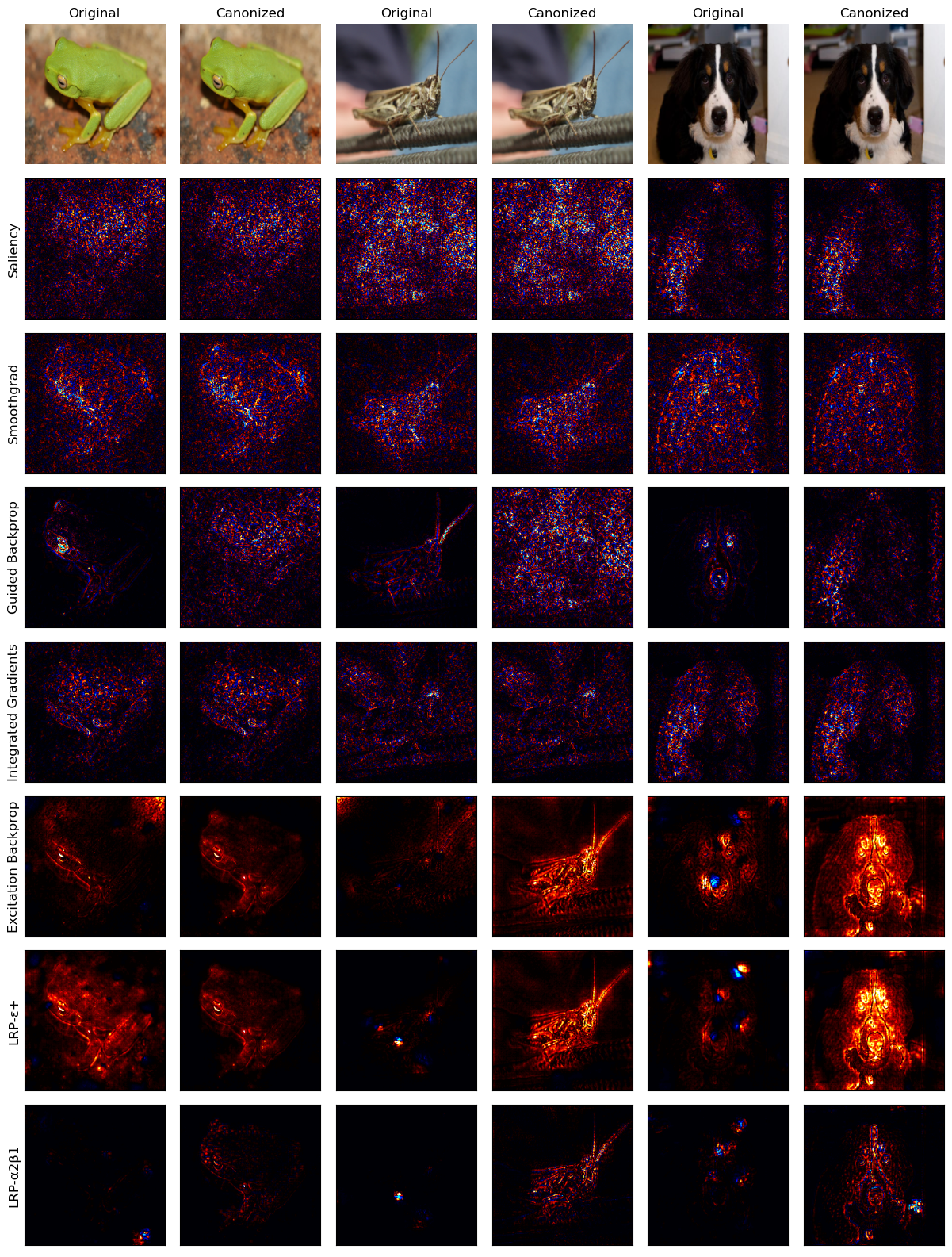}
  \caption{Attribution heatmaps with different XAI methods for Densenet-121 model on ILSVRC2017 dataset.}
  \label{fig:attr_densenet_121}
\end{figure*}

\begin{figure*}[ht]
    \centering
  \includegraphics[width=.8\linewidth]{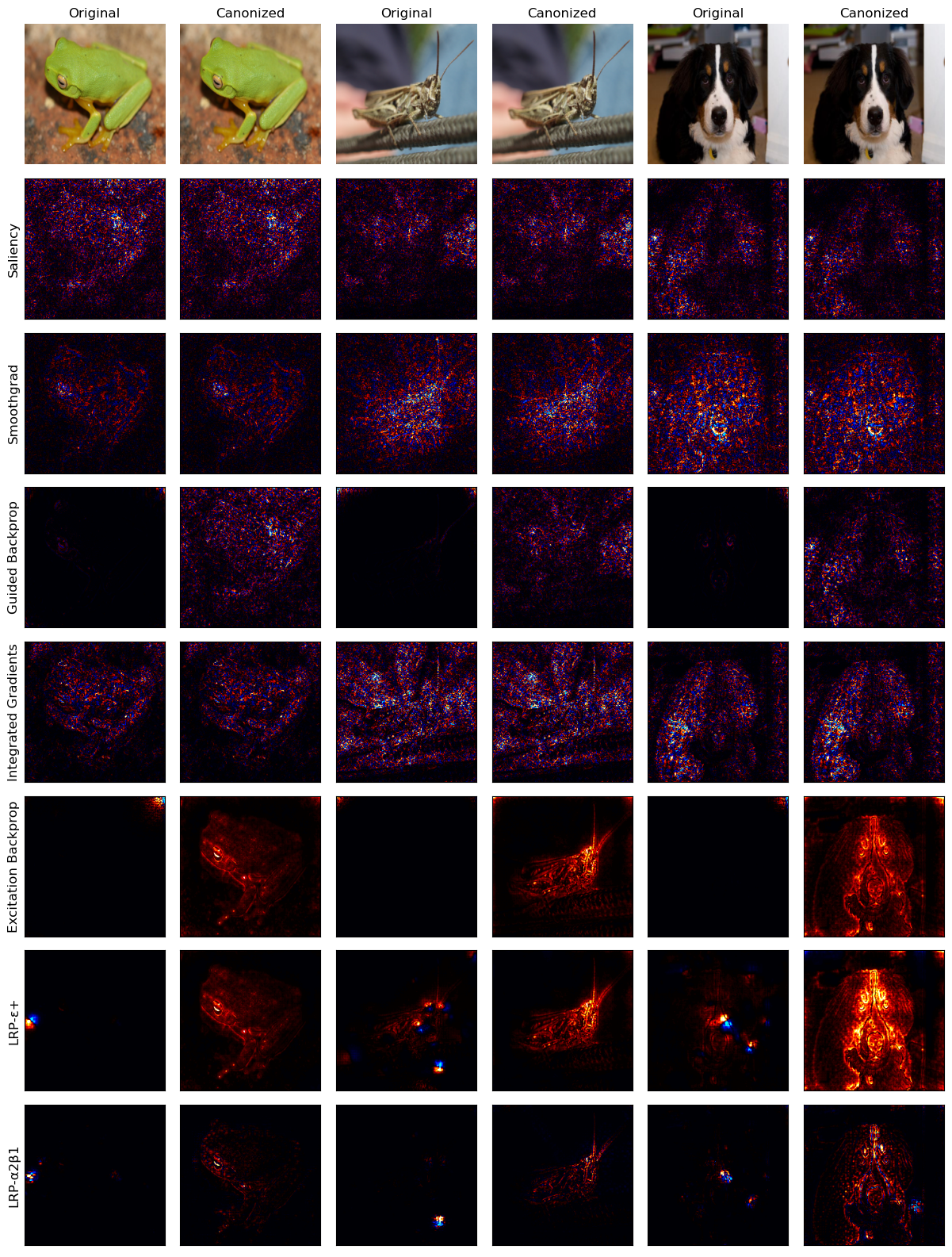}
  \caption{Attribution heatmaps with different XAI methods for Densenet-161 model on ILSVRC2017 dataset.}
  \label{fig:attr_densenet_161}
\end{figure*}

\begin{figure*}[ht]
    \centering
  \includegraphics[width=.75\linewidth]{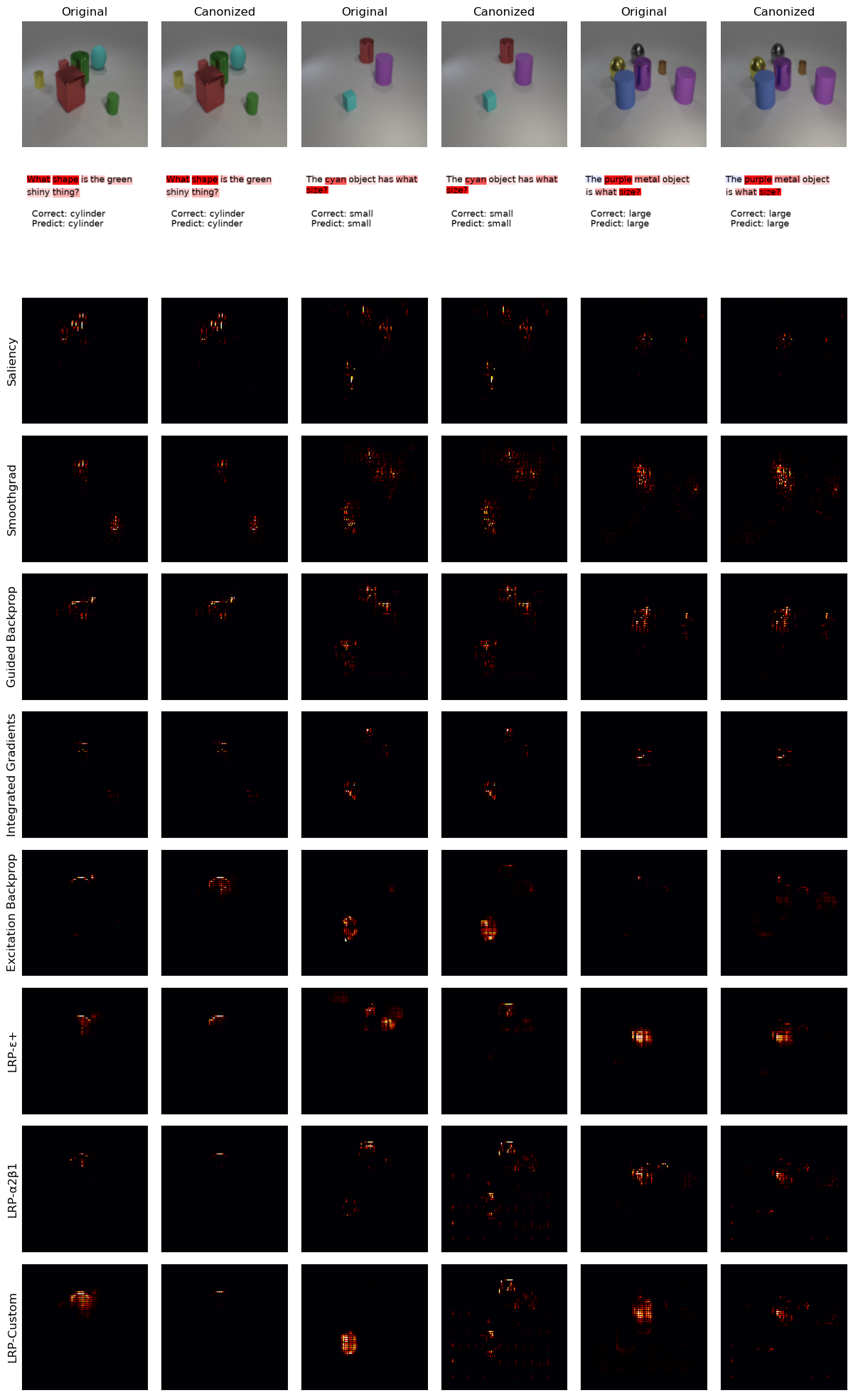}
  \caption{Attribution heatmaps for Relation Network on CLEVR-XAI dataset using \textit{pos-l2-norm-sq} pooling.}
  \label{fig:attr_clevr_pos_sq_sum}
\end{figure*}

\begin{figure*}[ht]
    \centering
  \includegraphics[width=.75\linewidth]{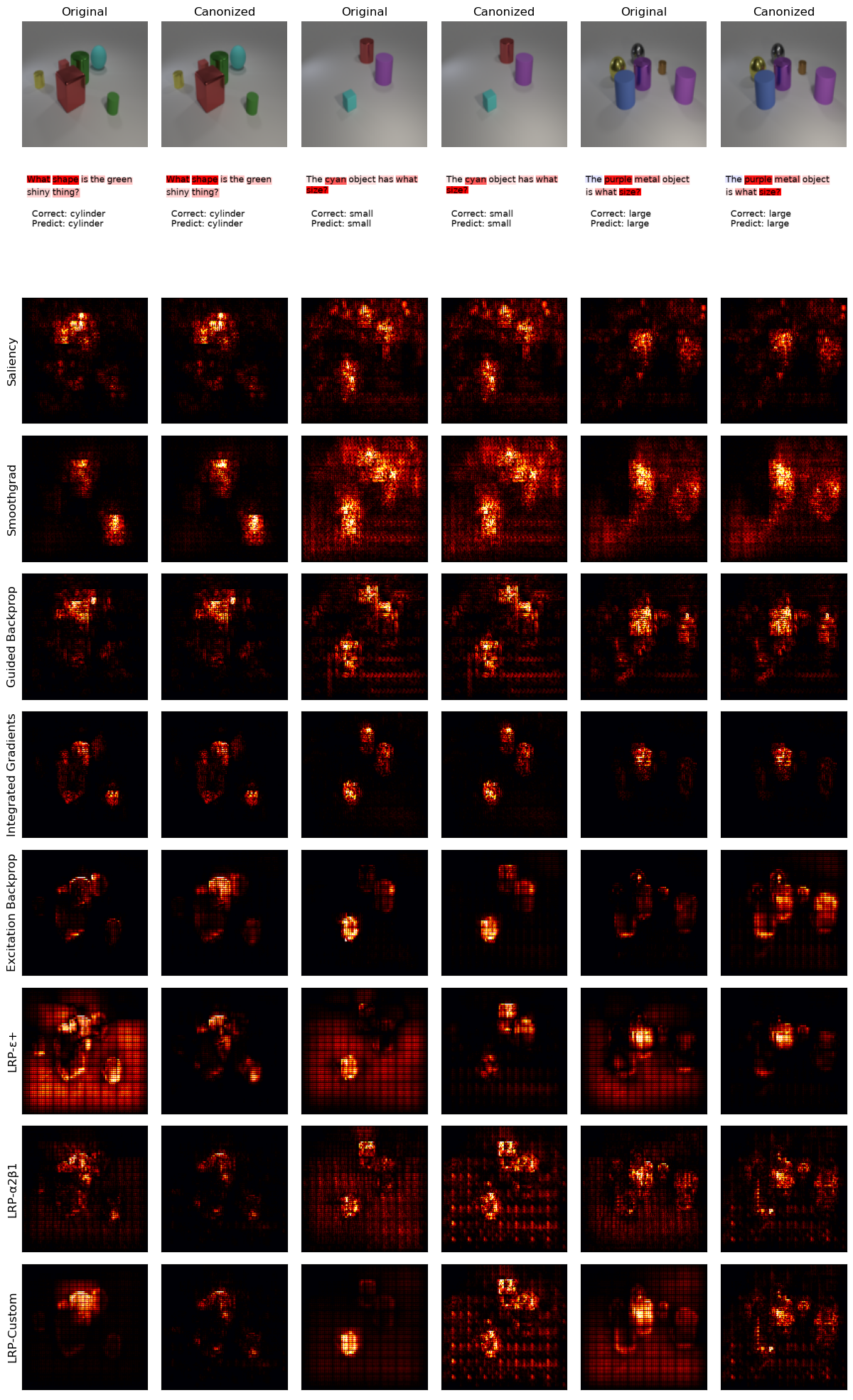}
  \caption{Attribution heatmaps for Relation Network on CLEVR-XAI dataset using \textit{max-norm} pooling.}
  \label{fig:attr_clevr_max_norm}
\end{figure*}

In Figures~\ref{fig:attr_vgg16}~-~\ref{fig:attr_densenet_161} we show attribution heatmaps for three samples using various XAI methods, both with and without model canonization using the ILSVRC2017 dataset for different model architectures.
Similarly, in Figures~\ref{fig:attr_clevr_pos_sq_sum}~-~\ref{fig:attr_clevr_max_norm} we show attribution heatmaps for different XAI methods with and without canonization for Relation Networks using \textit{pos-l2-norm-sq} pooling and \textit{max-norm} pooling.

\end{appendices}
\end{document}